\documentclass[Afour,sageh,times]{sagej}

\usepackage{moreverb,url}
\usepackage{marvosym}
\usepackage{array}
\usepackage{multirow}
\usepackage{threeparttable}
\usepackage{colortbl}
\usepackage[dvipsnames]{xcolor}
\usepackage{nameref}

\usepackage[colorlinks,bookmarksopen,bookmarksnumbered,citecolor=red,urlcolor=red]{hyperref}

\newcommand\BibTeX{{\rmfamily B\kern-.05em \textsc{i\kern-.025em b}\kern-.08em
T\kern-.1667em\lower.7ex\hbox{E}\kern-.125emX}}

\def\volumeyear{2016}

\begin{document}

\runninghead{SR-LIO++}

\title{SR-LIO++: LiDAR-Inertial Odometry and Quantized Mapping with Caching-Aware Sweep Reconstruction}

\author{Zikang~Yuan\affilnum{1,2}, Ruiye~Ming\affilnum{3}, Chengwei~Zhao\affilnum{4}, Yonghao~Tan\affilnum{1,2}, Pingcheng~Dong\affilnum{1,2}, Yuan~Ren\affilnum{5}, Yuzhong~Jiao\affilnum{1,2}, Xin~Yang\affilnum{3} and Kwang-Ting Cheng\affilnum{1,2}}

\affiliation{\affilnum{1}ACCESS-AI Chip Center for Emerging Smart Systems, InnoHK Centers, Hong Kong Science Park, Hong Kong, China\\
\affilnum{2}HongKong University of Science and Technology, HongKong, China\\
\affilnum{3}Huazhong University of Science and Technology, Wuhan, China\\
\affilnum{4}Hangzhou Qisheng Intelligent Techology Co. Ltd., Hangzhou, China\\
\affilnum{5}Southeast University, Nanjing, China}

\corrauth{Xin~Yang, School of Electronic Information and Communication Engineering,
Huazhong University of Science and Technology,
Wuhan City, Hubei Province, China.}

\email{xinyang2014@hust.edu.cn}

\begin{abstract}
	
Addressing the inherent low acquisition frequency limitation of 3D LiDAR to achieve high-frequency output has become a critical research focus in the LiDAR-Inertial Odometry (LIO) domain. To ensure real-time performance, frequency-enhanced LIO systems must process each sweep within significantly reduced timeframe, which presents substantial challenges for deployment on resource-constrained platforms. To address these limitations, we introduce SR-LIO++, an innovative LIO system capable of achieving doubled output frequency relative to input frequency on resource-constrained hardware platforms, including the Raspberry Pi 4B. Our system employs the previously proposed sweep reconstruction methodology to enhance LiDAR sweep frequency, generating high-frequency reconstructed sweeps. Building upon this foundation, we propose a caching mechanism for intermediate results (i.e., surface parameters) of the most recent segments, effectively minimizing redundant processing of common segments in adjacent reconstructed sweeps. This method decouples processing time from the traditionally linear dependence on reconstructed sweep frequency. Furthermore, we present a quantized map point management based on index table mapping, significantly reducing memory usage by converting global 3D point storage from 64-bit double precision to 8-bit char representation. This method also converts the computationally intensive Euclidean distance calculations in nearest neighbor searches from 64-bit double precision to 16-bit short and 32-bit integer formats, reducing computational cost. Extensive experimental evaluations across three distinct computing platforms and four public datasets demonstrate that SR-LIO++ maintains state-of-the-art accuracy while substantially enhancing efficiency. Notably, our system successfully achieves 20\,Hz state output on Raspberry Pi 4B hardware.

\end{abstract}


\maketitle

\section{Introduction}
\label{Introduction}

LiDAR-inertial odometry (LIO) have been widely recognized as a fundamental solution for localization and mapping in mobile robotics. Recent advancements have led to the development of numerous lightweight LIO systems [\cite{xu2021fast, xu2022fast, chen2023direct, chen2024ig, bai2022faster}], which demonstrate remarkable capabilities in achieving accurate and robust localization and mapping while maintaining exceptionally low computational overhead. The emergence of these efficient systems indicates a paradigm shift in LIO development, where the primary limiting factor for output frequency is no longer computational efficiency, but rather the inherent low data acquisition frequency of 3D LiDAR sensors. As discussed in Point-LIO [\cite{he2023point}], the limited output frequency will cause a delay in the odometry equal to a full sweep duration, and put an unnecessary upper bound for the odometry bandwidth due to the Nyquist–Shannon sampling theorem [\cite{qu2022llol}].

\begin{figure*}
	\setlength{\fboxsep}{0pt}%
	\setlength{\fboxrule}{0pt}%
	\begin{center}
		\includegraphics[width=0.97\textwidth]{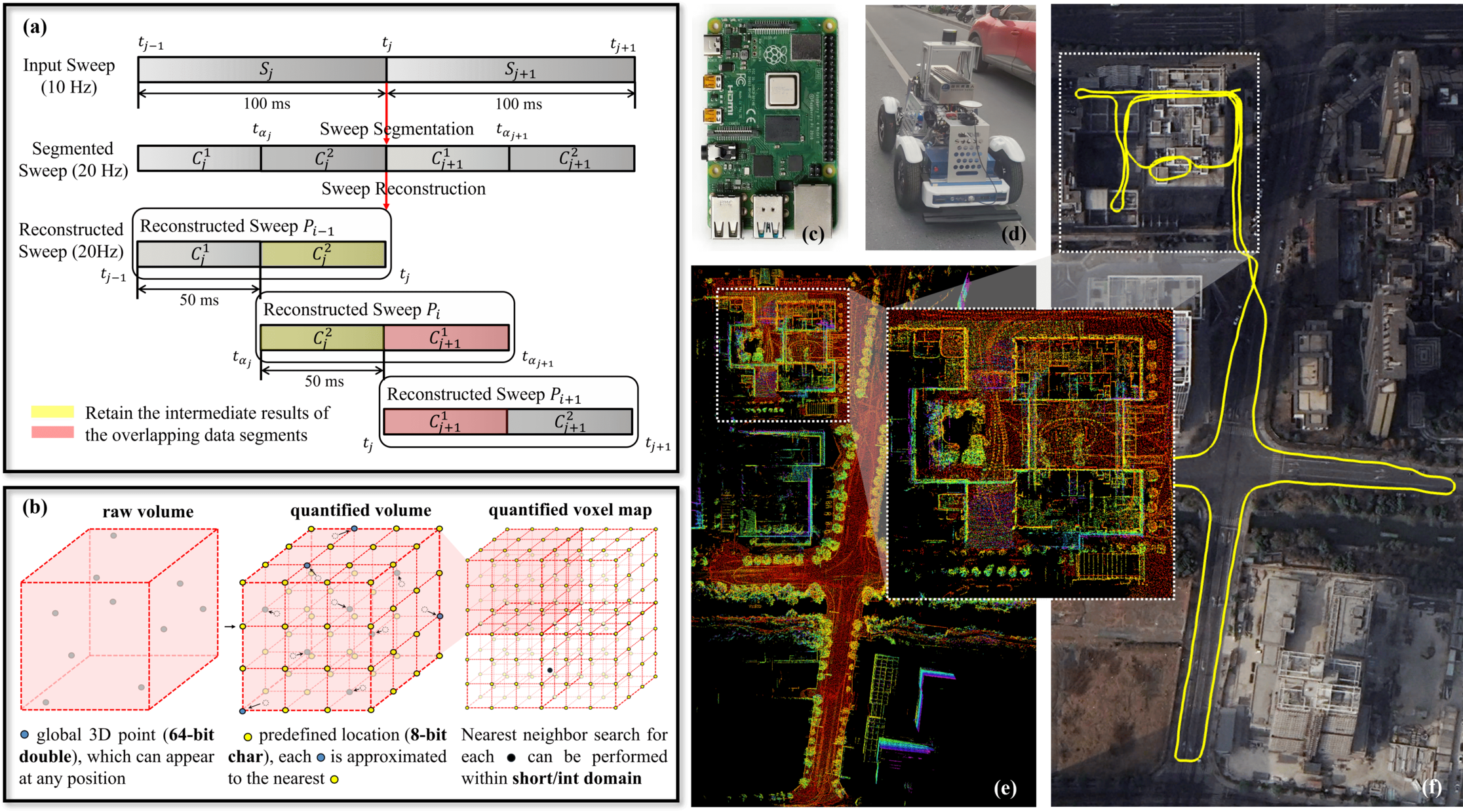}
	\end{center}
	\caption{(a) Illustration of the intermediate result (i.e., surface parameter) reutilization for overlapping segments in sweep reconstruction. The proposed method employs minimal storage overhead to eliminate redundant processing of overlapping segments across consecutive reconstructed sweeps. (b) Illustration of quantized map point management, which restrict global map points to predefined locations (as illustrated by yellow points), thereby enabling encoding 3D points with fewer bits and in turn optimizing computational efficiency and reducing memory overhead. (c) Benefiting from methods (a) and (b), SR-LIO++ is capable of real-time running at the frequency of 20\,Hz on Raspberry Pi 4B. (d) We deployed our own platform for data collection and testing in real world. The reconstructed 3D point cloud and the trajectory overlaid on Google Earth are shown in (e) and (f) respectively.}
	\label{fig1}
\end{figure*}

To address the above challenge, several frequency-enhanced LIO have been proposed, including the segmentation of 360-degree LiDAR sweeps into discrete segments [\cite{qu2022llol}] or even individual points [\cite{he2023point}]. They provides high-rate pose and map updates critical for stabilizing control loops and enabling rapid reaction to obstacles in high-speed autonomous robots. While these approaches effectively increase the LiDAR data acquisition frequency, they fundamentally alter the spatial distribution characteristics of the point cloud, compromising the inherent uniformity of 360-degree scan data and consequently reducing system accuracy and robustness. In contrast, our previous work SR-LIO [\cite{yuan2022sr}] introduces a novel sweep reconstruction method [\cite{yuan2023sdv, yuan2024sr}] (as illustrated in Fig. \ref{fig1}(a)), which achieves a 20\,Hz point cloud data frequency while maintaining the structural integrity of individual sweeps. Although SR-LIO demonstrates stable performance with doubled output frequency on advanced CPU architectures, its computational overhead increases linearly with the frequency of reconstructed sweeps, posing significant challenges for deployment on mobile robotic platforms. These platforms typically employ suboptimal CPU configurations, with many 3D laser scanners and robot vacuums utilizing ARM-based microcomputer motherboards. The limited computational resources of mobile platforms pose a fundamental barrier to implementing sweep reconstruction technique, as it cannot sustain the 20\,Hz processing frequency required by SR-LIO.

To enable universal and efficient deployment of sweep reconstruction on resource-constrained platforms, we present SR-LIO++ in this work. Our proposed system addresses two critical limitations of SR-LIO: 1) Redundant processing of overlapping data segments (illustrated by yellow and red segments in Fig. \ref{fig1} (a)) between adjacent reconstructed sweeps, which results in a linear increase in computational overhead as sweep segmentation frequency; and 2) The computational inefficiency of standard voxel map management, particularly in nearest neighbor searches which require repetitive Euclidean distance calculations using high-precision floating-point numbers (64-bit double type). To resolve the first limitation, we introduce a novel surface parameter reutilization method. This approach employs a minimal-overhead temporary cache to store intermediate results (surface parameters) of overlapping data segments, effectively eliminating redundant computations. Specifically, for a point $\mathbf{p} \in C_{j+1}^1 \subset P_i$ in Fig. \ref{fig1} (a), we compute its nearest neighbors from the global map once and utilize them to fit a surface for constructing point-to-plane residuals. The derived surface parameters (normal vector and normal offset) are then stored in the temporary cache. When subsequent reconstructed sweep $P_{i+1}$ arrives, these pre-stored surface parameters are directly retrieved, bypassing redundant computations for keypoints in segment $C_{j+1}^1$ and significantly enhancing computational efficiency. For the second limitation, we propose a novel quantized map point management method based on index table mapping. This method transforms the repetitive Euclidean distance calculations from 64-bit double precision to encoded 16-bit short and 32-bit integer formats. Specifically, our approach restricts global 3D points to predefined locations within the voxel map (illustrated by yellow points in Fig. \ref{fig1}(b)), rather than allowing arbitrary placement. This spatial constraint enables efficient encoding of 3D points utilizing low bit-depth. We encode the global map points with a 3.9\,mm resolution, such minor precision loss is fully acceptable for state estimation in urban scenarios. The encoded points not only minimize memory overhead but also replace floating-point Euclidean distance calculations with integer-based sum-of-squares operations during nearest neighbor searches, achieving substantial improvements in both memory efficiency and computational performance. We conducted extensive evaluations of SR-LIO++ across three computing platforms with varying computational capabilities, utilizing four public datasets [\cite{carlevaris2016university, yaneu, wen2020urbanloco, jeong2019complex}]. Experimental results demonstrate that our proposed methods maintain state-of-the-art accuracy while achieving 42.95$\%$ reduction in memory overhead and 22.45$\%$ improvement in computational efficiency on Raspberry Pi 4B. Notably, SR-LIO++ successfully achieves a 20\,Hz output frequency on Raspberry Pi 4B hardware. Furthermore, we validated our system's performance through real-world testing using collected data from our own platform, demonstrating its practical effectiveness in real urban environments.

To summarize, the main contributions of this work are four folds: 1) We introduce a novel surface parameter reutilization method that employs minimal storage overhead to maximally eliminate redundant processing of overlapping segments between adjacent reconstructed sweeps; 2) We develop a quantized map point management method based on index table mapping, which achieves significant improvements in both memory efficiency and computational performance; 3) We conduct comprehensive experimental evaluations to demonstrate that SR-LIO++ substantially reduces computational and storage overhead compared to SR-LIO while maintaining comparable accuracy and robustness. Additional validation using our proprietary dataset confirms the system's effectiveness in real-world scenarios; 4) To support community development and facilitate further research, we commit to immediately releasing the complete source code of SR-LIO++ upon acceptance of this manuscript.

The rest of this paper is structured as follows. In the Related Work section, we briefly discuss the relevant literature. The Preliminary section provides preparation contents. Then the Methodology section details our system SR-LIO++. The Experiments section provides experimental evaluation. Finally, we conclude the paper in the Conclusion section.

\section{Related Work}
\label{Related Work}

Over the past decade, LiDAR odometry (LO) and LiDAR-inertial odometry (LIO) have evolved through three distinct research phases. The initial phase focused on high-precision localization and robust mapping, establishing the foundational requirements for practical deployment. This success shifted attention to system optimization and lightweight implementation, leading to the development of computationally efficient yet reliable systems. More recently, the research frontier has advanced to overcome the inherent low acquisition frequency of 3D LiDAR sensors, which has long constrained the output rates of lightweight systems. This section systematically reviews the landmark works that have defined these three critical phases.

\subsection{Heavyweight LO/LIO}
\label{Heavyweight LO/LIO}

LOAM [\cite{zhang2014loam}], as the pioneering LO systems, establish a fundamental framework. Building upon LOAM, LeGO-LOAM [\cite{shan2018lego}] introduces a method to cluster raw point clouds and eliminate those with weak geometric structure information to reduce computational load. Nevertheless, accurately identifying and removing clusters with weak geometry is a nontrivial task, and improper removal of useful clusters can compromise the accuracy and robustness of pose estimation. SuMa [\cite{behley2018efficient}] proposes a surfel-based map representation that aggregates information from point clouds. However, real-time performance in SuMa necessitates GPU acceleration, and its pose estimation accuracy does not surpass that of LOAM-based systems. F-LOAM [\cite{wang2021}] streamlines the pose estimation pipeline by eliminating sweep-to-sweep estimation and employing analytic derivatives, while its extension ISC-LOAM [\cite{wang2020intensity}] incorporates intensity scan context for enhanced loop detection. IMLS-SLAM [\cite{deschaud2018imls}] proposes an Implicit Moving Least Squares (IMLS) algorithm as an alternative to ICP, though at significant computational cost. The integration of inertial measurements has led to notable advancements, including LINs [\cite{qin2020lins}], which implements an error state iterated kalman filter (ESIKF) framework for LiDAR-IMU fusion, and LIO-SAM [\cite{shan2020lio}], which pioneers factor graph optimization in LIO systems. Recent trends have focused on optimization frameworks and computational efficiency. \cite{ye2019tightly} introduces a Bundle Adjustment (BA) framework for LiDAR-IMU fusion, complemented by rotation-constrained refinement. \cite{li2021towards} extends this approach with key-sweep selection and multi-sweep optimization, while LIO-Livox [\cite{livox2021}] provides an open-source framework with dynamic point removal and feature-based optimization. \cite{wang2021lightweight} enhances real-time performance through point-to-plane constraints and IMU pre-integration. The emergence of elastic odometry frameworks, including CT-ICP [\cite{dellenbach2022ct}] with its dual-state optimization and logical constraints, and \cite{yuan2025semi} with the semi-elastic optimization method, represent significant advancements in state estimation and distortion calibration, albeit with ongoing challenges in computational efficiency. Furthermore, there have been very recent developments aimed at enhancing robustness and reducing hyperparameter tuning. These efforts include not only the minimalist approach [\cite{vizzo2023kiss}], but also the introduction of novel correspondence mechanisms [\cite{ferrari2024mad, lee2024genz}].

\subsection{Lightweight LO/LIO}
\label{Lightweight LO/LIO}

The advent of Fast-LIO [\cite{xu2021fast}] represents a seminal milestone in the evolution of LiDAR-SLAM, marking the transition to an era dominated by computationally efficient solutions. This pioneering work introduces an innovative Kalman gain computation technique [\cite{sorenson1966kalman}] that circumvents the need for high-order matrix inversion, substantially reducing computational complexity. Building upon this foundation, Fast-LIO2 [\cite{xu2022fast}] presents the ikd-tree [\cite{cai2021ikd}] algorithm, which significantly optimizes tree construction, traversal, and element removal operations compared to conventional kd-tree implementations. Further advancements include Faster-LIO [\cite{bai2022faster}], which introduces an incremental voxel-based approach capable of efficiently processing both spinning and solid-state LiDAR data through parallelized approximate k-nearest neighbor (kNN) queries. Voxel-Map [\cite{yuan2022efficient}] introduces a hierarchical voxel mapping system utilizing Hash tables and octrees, enabling efficient map construction and updates through a probabilistic adaptive framework. DLIO [\cite{chen2023direct}] implements a third-order minimal preservation strategy for state prediction and point distortion calibration, achieving superior pose estimation accuracy. IG-LIO [\cite{chen2024ig}] integrates generalized-ICP (GICP) constraints with inertial measurements within a unified estimation framework, complemented by a voxel-based surface covariance estimator for probabilistic environment modeling. The emergence of multi-LiDAR systems is represented by \cite{chen2024multi}, which employs parallel state updates and voxelized map representations to efficiently process data from multiple LiDAR sensors. \cite{yuan2025dynamic} employs a voxel-based map management strategy and an ESIKF-based state estimation approach, integrated with a label consistency dynamic object removal scheme to achieve a lightweight odometry framework for dynamic urban scenes. These advancements collectively demonstrate that contemporary LO/LIO systems can process individual sweeps with minimal computational overhead, shifting the fundamental limitation of system performance from computational efficiency to the inherent low acquisition frequency of 3D LiDAR sensors.

\subsection{Frequency-Enhanced LO/LIO}
\label{LO/LIO with frequency increasement}

To overcome the inherent limitation of low acquisition frequency in 3D LiDAR systems, several innovative approaches have been proposed. LLOL [\cite{qu2022llol}] pioneers a streaming sensor paradigm, processing LiDAR data packets incrementally upon arrival. This distributed processing approach significantly enhances system throughput while reducing latency, resulting in a highly efficient and lightweight architecture. Building upon this concept, Point-LIO [\cite{he2023point}] introduces a point-by-point processing framework that enables state updates at the frequency of individual LiDAR point measurements, achieving unprecedented output frequencies. However, these methods fundamentally alter the spatial distribution characteristics of the point cloud, compromising the inherent 360-degree uniformity of LiDAR sweeps and consequently reducing system accuracy and robustness. Alternative approaches have focused on maintaining data integrity while increasing processing frequency. SDV-LOAM [\cite{yuan2023sdv}] introduces a novel sweep reconstruction method that combines current segments with historical data to generate complete 360-degree sweeps, effectively increasing point cloud frequency while preserving spatial integrity. Despite its innovative approach, SDV-LOAM's complex framework based on DSO [\cite{engel2017direct}] prevents real-time operation. Our previous work SR-LIO [\cite{yuan2022sr}] addresses this limitation by implementing sweep reconstruction within a streamlined ESIKF framework, achieving a 2X frequency enhancement on advanced CPU. Subsequent developments, including \cite{zhang2024lio} and \cite{huang2024lio}, have further demonstrated the effectiveness of sweep reconstruction in mitigating point cloud distortion under aggressive motion conditions. While these advancements have proven effective on personal computing platforms, their deployment on mobile robotic systems presents significant challenges. Mobile platforms typically employ less powerful CPUs compared to desktop systems, with many 3D laser scanners and robot vacuums utilizing ARM-based microcomputer architectures. These inherent computational limitations create substantial barriers to implementing sweep reconstruction methods in mobile applications.

\section{Preliminary}
\label{Preliminary}

\subsection{Notation}
\label{Notation}

\begin{table}[t]
	\small\sf\centering
	\caption{Definition of generalized variation.\label{table1}}
	\vspace{-0.2cm}
	\begin{center}
	\begin{threeparttable}
		\begin{tabular}{>{\centering\arraybackslash}p{0.15cm} >{\centering\arraybackslash}p{3.1cm} >{\raggedright\arraybackslash}p{3.9cm}}
			\toprule
			& \centering Type of Variables & Computational Formula \\ \hline
			\multirow{4}{*}{$\boxplus$} & $\mathbf{a}, \mathbf{b} \in \mathbb{R}^3$ & $\mathbf{a} \boxplus \mathbf{b} = \mathbf{a} + \mathbf{b}$ \\
			& $\mathbf{R} \in SO(3)$, $\boldsymbol{\theta} \in so(3)$ & $\mathbf{R} \boxplus \boldsymbol{\theta} = \mathbf{R} \mathrm{Exp}(\boldsymbol{\theta})$ \\
			& \multirow{2}{*}{$\mathbf{g}_1, \mathbf{g}_2 \in \mathbb{R}^3$, $\delta \mathbf{g} \in S^2$} & $\mathbf{g}_2 = \mathbf{g}_1 \boxplus \delta \mathbf{g} = \mathrm{Exp}(\mathbf{B}(\mathbf{g}_1) \delta \mathbf{g}) \mathbf{g}_1$ \\ \hline
			\multirow{3}{*}{$\boxminus$} & $\mathbf{a}, \mathbf{b} \in \mathbb{R}^3$ & $\mathbf{a} \boxminus \mathbf{b}=\mathbf{a}-\mathbf{b}$ \\
			& $\mathbf{R}_1, \mathbf{R}_2 \in SO(3)$ & $\mathbf{R}_1 \boxminus \mathbf{R}_2 = \mathrm{Log}({\mathbf{R}_2}^T\mathbf{R}_1)$ \\
			& $\mathbf{g}_1, \mathbf{g}_2 \in \mathbb{R}^3$, $\delta \mathbf{g} \in S^2$ & $\delta \mathbf{g} = \mathbf{g}_1 \boxminus \mathbf{g}_2 = \mathbf{B}{(\mathbf{g}_2)}^T \delta \boldsymbol{\theta}_{\mathbf{g}}$ \\ \bottomrule
		\end{tabular}
	\end{threeparttable}
	\begin{tablenotes}
		\footnotesize
		\item[] \textbf{Denotations}: $\boldsymbol{\theta}_{\mathbf{g}}$ is the Lie algebra of rotation from $\mathbf{g}_1$ to $\mathbf{g}_2$.
	\end{tablenotes}
	\end{center}
\end{table}

We define $(\cdot)^w$, $(\cdot)^l$ and $(\cdot)^b$ as the representations of a 3D point in the world coordinate system, LiDAR coordinate system, and IMU coordinate system, respectively. The world coordinate system is initialized to coincide with the IMU coordinate system $(\cdot)^b$ at the starting position. For the $i$-th LiDAR sweep acquired at time $t_i$, the pose transformation from the IMU coordinate system $(\cdot)^{b_i}$ to the world coordinate system $(\cdot)^{w}$ is precisely defined as $\mathbf{T}_{b_i}^{w}$, comprising a rotation matrix $\mathbf{R}_{b_i}^{w} \in S O(3)$ and a translation vector $\mathbf{t}_{b_i}^{w} \in \mathbb{R}^3$. In addition to the pose estimation, our state vector incorporates several crucial parameters: the velocity $\mathbf{v}$, accelerometer bias $\mathbf{b}_{\mathbf{a}}$, gyroscope bias $\mathbf{b}_{\boldsymbol{\omega}}$ and gravitational acceleration $\mathbf{g}^{w}$, providing a comprehensive representation by a vector:
{\small
\begin{equation}
	\label{equation1}
	\boldsymbol{x}=\left[\mathbf{t}^T, \mathbf{q}^T, \mathbf{v}^T, {\mathbf{b}_{\mathbf{a}}}^T, {\mathbf{b}_{\boldsymbol{\omega}}}^T, {\mathbf{g}^{w}}^T \right]^T
\end{equation}
}
where $\mathbf{q}$ is the quaternion form of the rotation matrix $\mathbf{R}$. Accordingly, the error state is expressed as:
{\small
\begin{equation}
	\label{equation2}
	\delta \boldsymbol{x}=\left[\delta \mathbf{t}, \delta \boldsymbol{\theta}, \delta \mathbf{v}, \delta \mathbf{b}_{\mathbf{a}}, \delta \mathbf{b}_{\boldsymbol{\omega}}, \delta \mathbf{g}\right]^T
\end{equation}
}
It is necessary to note that $\delta \boldsymbol{\theta} \in so(3)$, which is the Lie algebra of rotation. $\delta \mathbf{t}$, $\delta \mathbf{v}$, $\mathbf{b}_{\mathbf{a}}$, $\delta \mathbf{b}_{\boldsymbol{\omega}} \in \mathbb{R}^3$, $\delta \mathbf{g} \in  S^2$ due to the fixed length of gravitational acceleration. When updating the nominal state with the error state, the linear addition cannot be directly applied. The same as \cite{xu2022fast}, we define generalized addition $\boxplus$ and generalized subtraction $\boxminus$ for iterative state update, which is defined in Table \ref{table1}. In this context, the definition of the matrix $\mathbf{B}(\mathbf{g})$ is as follows:
{\small
\begin{equation}
	\label{equation3}
	\mathbf{B}(\mathbf{g})=\left[\begin{array}{cc}
		1-\frac{\overline{\mathbf{g}}_x^2}{1+\overline{\mathbf{g}}_z} & -\frac{\overline{\mathbf{g}}_x \overline{\mathbf{g}}_y}{1+\overline{\mathbf{g}}_z} \\
		-\frac{\overline{\mathbf{g}}_x \overline{\mathbf{g}}_y}{1+\overline{\mathbf{g}}_z} & 1-\frac{\overline{\mathbf{g}}_y{ }^2}{1+\overline{\mathbf{g}}_z} \\
		-\overline{\mathbf{g}}_x & -\overline{\mathbf{g}}_y
	\end{array}\right]
\end{equation}
}
where $(\overline{\cdot})$ indicates normalization for the element in the bracket.

\subsection{Voxel Map Management}
\label{Voxel Map Management}

\begin{figure}
	\setlength{\fboxsep}{0pt}%
	\setlength{\fboxrule}{0pt}%
	\begin{center}
		\includegraphics[width=0.95\linewidth]{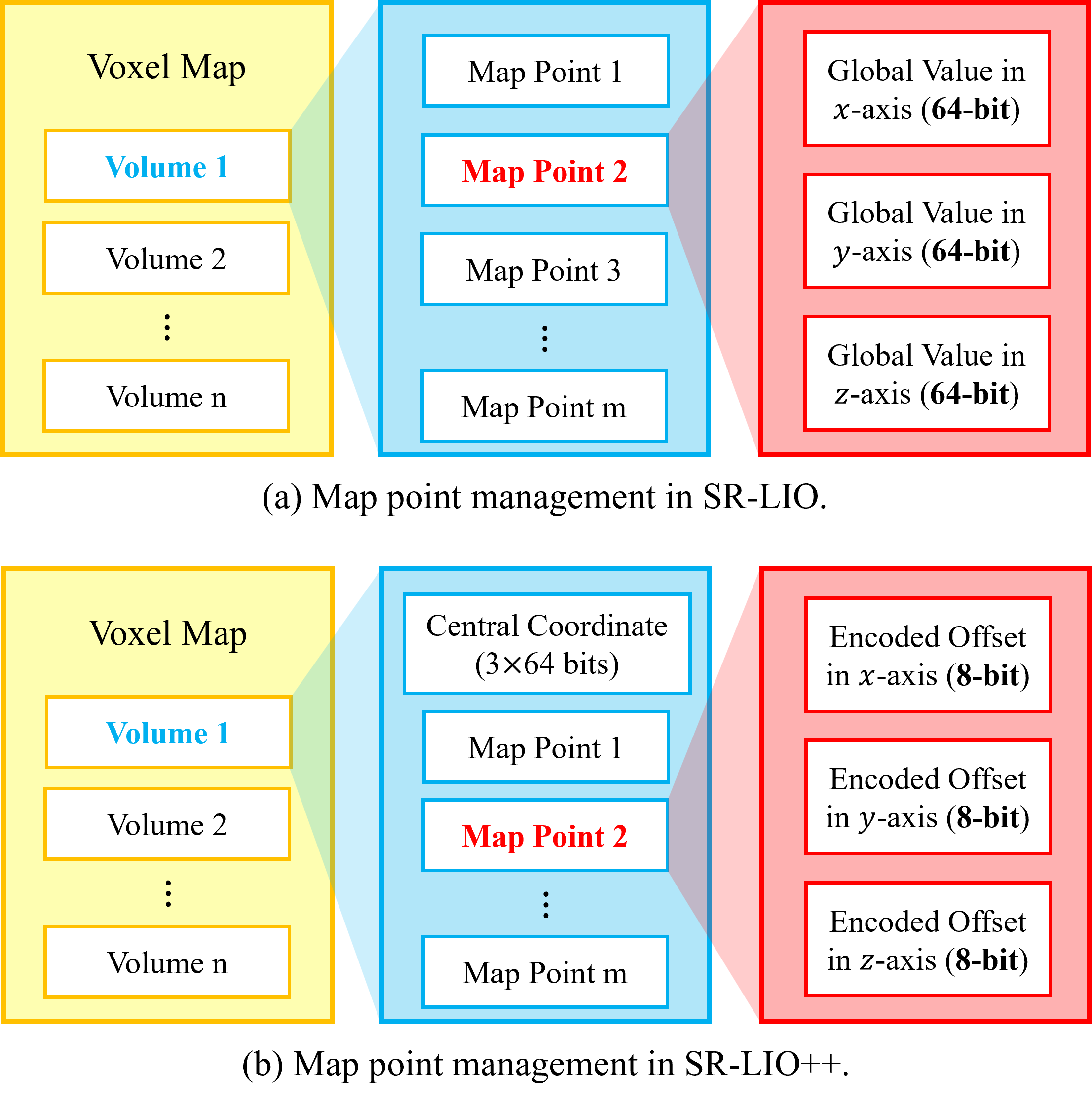}
	\end{center}
		\caption{Comparison of the map point management of SR-LIO and SR-LIO++.}
		\label{fig2}
\end{figure}

The system maintains a global map organized through a voxel-based data structure. As depicted in Fig. \ref{fig2} (a), the voxel map in SR-LIO [\cite{yuan2022sr}] comprises multiple volumetric units, with the number of units expanding as the robotic platform explores previously unmapped environments. Each volumetric unit measures 1.0$\times$1.0$\times$1.0 (unit: m) and contains a maximum of 20 points. Each point is represented by its global coordinates ($x$, $y$, $z$), with each coordinate stored as a 64-bit value.

In SR-LIO++, we implement an optimized storage scheme where each volumetric unit is characterized by a central coordinate stored as three 64-bit variables. Subsequently, individual 3D points within each volume are encoded as offsets relative to this central coordinate. Given the bounded range of the bias values (-0.5 to 0.5\,m), we can encode them as 8-bit integers. The specific quantization encoding scheme will be detailed in the Quantized Map Point Management section.

\subsection{Parameter Configuration}
\label{Parameter Configuration}

\begin{table}[t]
	\small\sf\centering
	\caption{Parameter configuration.\label{table2}}
	\vspace{-0.2cm}
	\begin{center}
		\begin{tabular}{>{\centering\arraybackslash}p{6.2cm} >{\centering\arraybackslash}p{1.0cm}}
			\toprule
			Parameter                            & Value  \\ \hline
			\texttt{num of selected points per sweep} & 600    \\
			\texttt{maximum num of iterations}         & 5      \\
			\texttt{num of neighborhood volumes}       & 27     \\
			\texttt{num of nearest neighbors}          & 20     \\
			\texttt{size of each volume}                  & 1.0\,m \\
			\texttt{max num of points in a volume} & 20     \\ \bottomrule
		\end{tabular}
	\end{center}
\end{table}

The Iterative State Update section incorporates multiple configuration parameters that govern system performance and computational efficiency. These parameters encompass: 1) the number of selected keypoints per sweep; 2) the maximum iteration count; 3) the number of volumes traversed during nearest neighbor search; 4) the number of nearest neighbors employed for surface fitting; 5) the size of each volume; and 6) the maximum point capacity per volume. While conventional academic practice typically represents such parameters using symbolic notation, we have opted to present specific numerical values to facilitate intuitive understanding of computational overhead reduction at a fundamental level of our method. Each parameter is accompanied by appropriate prefixes or annotations to ensure clarity in interpretation. The parameter values, originally established by our previous work SR-LIO [\cite{yuan2022sr}], are systematically presented in Table \ref{table2}. Extensive empirical validation in SR-LIO has confirmed the robustness and general applicability of this parameter set across diverse operational scenarios.

\section{Methodology}
\label{Methodology}

\subsection{System Overview}
\label{System Overview}

\begin{figure*}
	\setlength{\fboxsep}{0pt}%
	\setlength{\fboxrule}{0pt}%
	\begin{center}
		\includegraphics[width=0.97\textwidth]{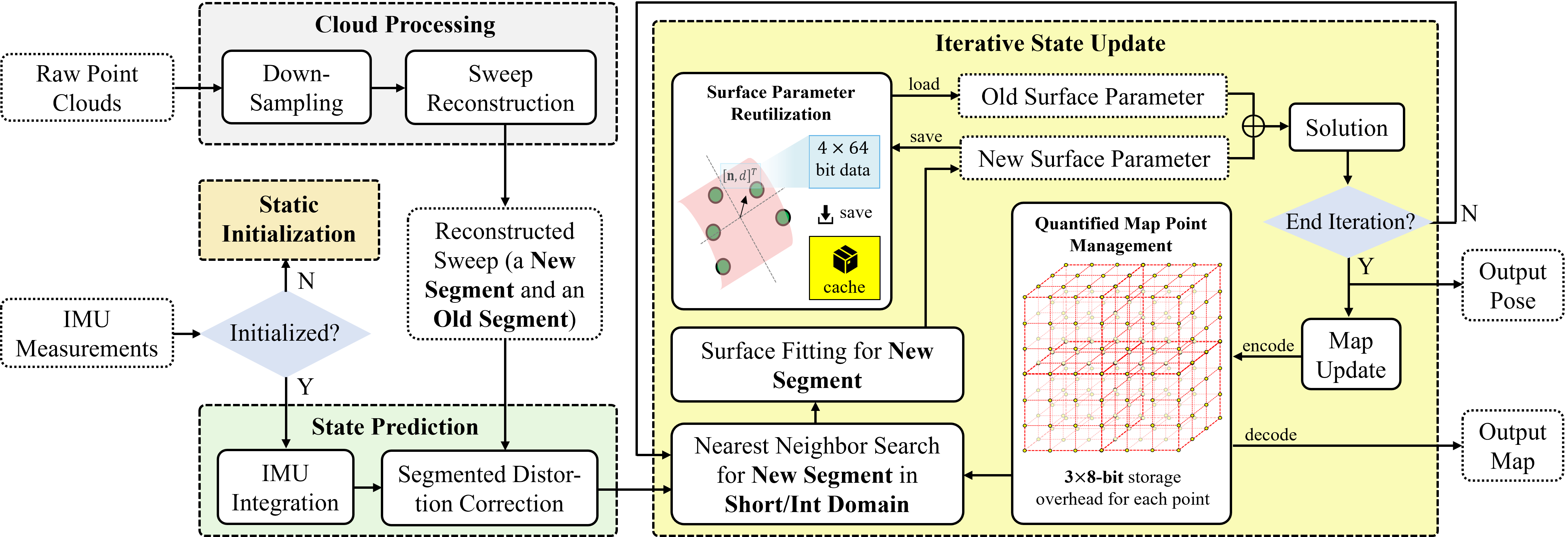}
	\end{center}
	\caption{System Overview of SR-LIO++. The overall system consists of a cloud processing module, a static initialization module, a state prediction module and an iterative state update module. The surface parameter reutilization and quantified map point management are two core contributions, where the former can significantly reduce computational overhead and the latter can reduce both computational and memory overhead simultaneously.}
	\label{fig3}
\end{figure*}

Fig. \ref{fig3} illustrates the framework of our SR-LIO++ which consists of four main modules: cloud processing, static initialization, state prediction and iterative state update. The cloud processing module down-samples the 10\,Hz input sweep, then segments and reconstructs the 10\,Hz down-sampled sweep to obtain 20\,Hz reconstructed sweep, while each reconstructed sweep consists of two segments. The static initialization module utilizes the IMU measurements to estimate some state parameters such as gravitational acceleration, accelerometer bias, gyroscope bias, and initial velocity. The state prediction module estimates an initial state using IMU raw measurements, and performs segmented distortion calibration for each reconstructed sweep. The iterative state update module first searches nearest neighbors for each keypoint of a new segment in encoded offset domain, then utilizes these neighbors fitting a surface to obtain the surface parameters. Next, the new surface parameters together with the previously stored old surface parameters, are used to construct point-to-plane constraints and solve current state. The entire process is iteratively executed until the convergence criterion is met or the maximum number of iterations is reached. Finally, upon termination of the iteration, we perform quantization on the new points, and subsequently incorporate the quantized points into the global voxel map.

\subsection{Cloud Processing}
\label{Cloud Processing}

\subsubsection{Down-Sampling}
\label{Down-Sampling}

To mitigate the substantial computational burden associated with processing a large volume of 3D point cloud data, we implement a down-sampling strategy to reduce the data volume. Initially, we apply a quantitative down-sampling approach, retaining one point for every four points in the original dataset. Subsequently, the down-sampled points are distributed into a volumetric grid with cell dimensions of 0.5$\times$0.5$\times$0.5 (unit: m), ensuring that each voxel contains no more than a single point.

\subsubsection{Sweep Reconstruction}
\label{Sweep Reconstruction}

The sweep reconstruction method was initially proposed in SDV-LOAM [\cite{yuan2023sdv}] and subsequently successfully implemented in SR-LIO [\cite{yuan2022sr}]. It represents a methodology for generating reconstructed sweeps at 20\,Hz from original 10\,Hz input sweeps. While the theoretical framework of sweep reconstruction permits frequency enhancement beyond 2X, SR-LIO's implementation is constrained to a twofold increase to maintain real-time processing capabilities and ensure accuracy within its state estimation module. The fundamental principle of sweep reconstruction has been detailed in SR-LIO [\cite{yuan2022sr}]. Different from SR-LIO, we do not regard the reconstructed sweep as a monolithic element. Instead, it is articulated into two distinct segments: a new segment and an old segment. The subsequent iterative state updates for these two segments will be processed in a differentiated manner.

\subsection{State Initialization}
\label{State Initialization}

In our system implementation, we employ a static initialization approach, as proposed in Open-VINs [\cite{geneva2020openvins}], to estimate several critical parameters, including the initial velocity vector, gravitational acceleration vector, and the biases of both the accelerometer and gyroscope sensors. For a comprehensive theoretical analysis and implementation details of the static initialization algorithm, readers are referred to the original work presented in Open-VINs [\cite{geneva2020openvins}].

\subsection{State Prediction}
\label{State Prediction}

\subsubsection{IMU Integration}
\label{IMU Integration}

The state prediction is performed once receiving an IMU input (i.e., $\hat{\boldsymbol{\omega}}_{n+1}$ and $\hat{\mathbf{a}}_{n+1}$), while the optimal state $\boldsymbol{x}_{n+1}^w$ (i.e., $\mathbf{t}_{n+1}^w$, $\mathbf{R}_{n+1}^w$, $\mathbf{v}_{n+1}^w$, $\mathbf{b}_{\mathbf{a}_{n+1}}$, $\mathbf{b}_{\boldsymbol{\omega}_{n+1}}$, $\mathbf{g}_{n+1}^w$) is calculated by:
{\small
\begin{equation}
	\label{equation6-1}
	\mathbf{R}_{n+1}^w=\mathbf{R}_n^w Exp\left(\left(\frac{\hat{\boldsymbol{\omega}}_n+\hat{\boldsymbol{\omega}}_{n+1}}{2}-\mathbf{b}_{\boldsymbol{\omega}_n}\right) \Delta t\right)
\end{equation}
}
{\small
\begin{equation}
	\label{equation6-2}
	\mathbf{v}_{n+1}^w=\mathbf{v}_n^w+\mathbf{R}_n^w\left(\frac{\hat{\mathbf{a}}_n+\hat{\mathbf{a}}_{n+1}}{2}-\mathbf{b}_{\mathbf{a}_n}-\mathbf{R}_w^n \mathbf{g}_n^w\right) \Delta t
\end{equation}
}
{\small
\begin{equation}
	\label{equation6-3}
	\mathbf{t}_{n+1}^w =\mathbf{t}_n^w+\mathbf{v}_n^w \Delta t+ \\ \frac{1}{2}\mathbf{R}_n^w\left(\frac{\hat{\mathbf{a}}_n+\hat{\mathbf{a}}_{n+1}}{2}-\mathbf{b}_{\mathbf{a}_n}- \mathbf{R}_w^n \mathbf{g}_n^w\right) \Delta t^2
\end{equation}
}
{\small
\begin{equation}
	\label{equation6-4}
	\mathbf{b}_{\mathbf{a}_{n+1}}=\mathbf{b}_{\mathbf{a}_n}, \mathbf{b}_{\boldsymbol{\omega}_{n+1}}=\mathbf{b}_{\boldsymbol{\omega}_n}, \mathbf{g}_{n+1}^w=\mathbf{g}_n^w
\end{equation}
}
The error state $\delta \boldsymbol{x}_{n+1}$ and covariance $\mathbf{\Sigma}_{n+1}$ is propagated as:
{\small
\begin{equation}
	\label{equation7-1}
	\delta \boldsymbol{x}_{n+1}=\mathbf{F}_{\boldsymbol{x}} \delta \boldsymbol{x}_n
\end{equation}
}
{\small
\begin{equation}
	\label{equation7-2}
	\mathbf{\Sigma}_{n+1}=\mathbf{F}_{\boldsymbol{x}} \mathbf{\Sigma}_n {\mathbf{F}_{\boldsymbol{x}}}^T+\mathbf{F}_{\boldsymbol{w}} \mathbf{Q} {\mathbf{F}_{\boldsymbol{w}}}^T
\end{equation}
}
where $\Delta t$ is the time interval between two consecutive IMU measurements. The formula of state transition matrix $\mathbf{F}_{\boldsymbol{x}}$, noise covariance matrix $\mathbf{Q}$ and noise Jacobian matrix $\mathbf{F}_{\boldsymbol{w}}$ have been detailed in SR-LIO [\cite{yuan2022sr}].

\subsubsection{Segmented Distortion Correction}
\label{Segmented Distortion Correction}

When applying conventional distortion calibration methods directly to adjacent reconstructed sweeps, which share overlapping data segments, the same portion of data is calibrated twice, leading to inconsistencies in the final results. We use the segmented distortion correction methodology for distortion correction of point cloud data, ensuring consistent and accurate point cloud representation under the usage of sweep reconstruction. The fundamental principle of segmented distortion correction has been detailed in SR-LIO [\cite{yuan2022sr}].

\subsection{Iterative State Update}
\label{Iterative State Update}

The iterative update of the error state through LiDAR point-to-plane constraints represents the core computational component within the ESIKF framework and constitutes the primary computational burden of the entire system. In SR-LIO's implementation, this process involves the random selection of a fixed number $k$ of keypoints ($k=600$) from the reconstructed sweep to form a keypoint set $K$ for each iteration, followed by the execution of the following four sequential steps:

\textbf{1) Point Transformation}: Each LiDAR point $\mathbf{p}_k \in K$ is transformed into the world coordinate system $(\cdot)^w$ using the current nominal state estimate. \textbf{2) Nearest Neighbor Search}: For each transformed point $\mathbf{p}_k^w$, the algorithm identifies its 20 nearest neighbors within the global map. \textbf{3) Surface Fitting}: A local planar surface is estimated from these neighboring points, characterized by a 3D normal vector $\mathbf{n}$ and a scalar offset $d$. \textbf{4) Error State Solution}: Point-to-plane residuals are constructed using the keypoint and the estimated plane parameters to compute the error state increment.

This iterative process continues until either the maximum iteration count is reached or the state increment converges below a specified threshold. Through comprehensive analysis, we have identified several redundant computational steps in SR-LIO's implementation that present opportunities for optimization. The subsequent sections of this chapter will elaborate on our proposed optimization approach and provide detailed computational analysis.

\subsubsection{Surface Parameter Reutilization}
\label{Surface Parameter Reutilization}

The computational process for each step is mathematically formulated and analyzed as follows: For step 1), the point transformation is mathematically expressed as:
{\small
\begin{equation}
	\label{equation15}
	\mathbf{p}_k^w=\mathbf{R}_{b_{i+1}}^w \mathbf{p}_k+\mathbf{t}_{b_{i+1}}^w
\end{equation}
}
where $R_{b_{i+1}}^w$ and $t_{b_{i+1}}^w$ represent the rotation matrix and translation vector from the body frame $(\cdot)^{b_{i+1}}$ to the world frame $(\cdot)^w$ at timestamp $t_{i+1}$, respectively. This transformation requires $k\times$$m$ matrix-vector multiplications and vector additions, where $m$ denotes the actual iteration count ranging from 1 to 5. In step 2), the algorithm performs nearest neighbor search by computing Euclidean distances between $\mathbf{p}_k^w$ and all points within the 27-neighborhood voxels of its residing volume. As established in the Voxel Map Management section, each voxel maintains a maximum of 20 points. Consequently, this step involves 27$\times$20$\times$$k$$\times$$m$ sum-of-squares operations for distance calculations and $k\times$$m$ sorting operations of size 27$\times$20. Step 3) implements surface parameter estimation through eigenvalue decomposition of the covariance matrix constructed from the 20 neighboring points identified in step 2). This computation is performed for each keypoint's neighborhood, resulting in $k\times$$m$ eigenvalue decompositions. Following the completion of steps 1) through 3), the system utilizes the $k$ transformed keypoints and their corresponding surface parameters to establish point-to-plane constraints, which are subsequently incorporated into the solution.

\begin{figure}
	\setlength{\fboxsep}{0pt}%
	\setlength{\fboxrule}{0pt}%
	\begin{center}
		\includegraphics[width=0.97\linewidth]{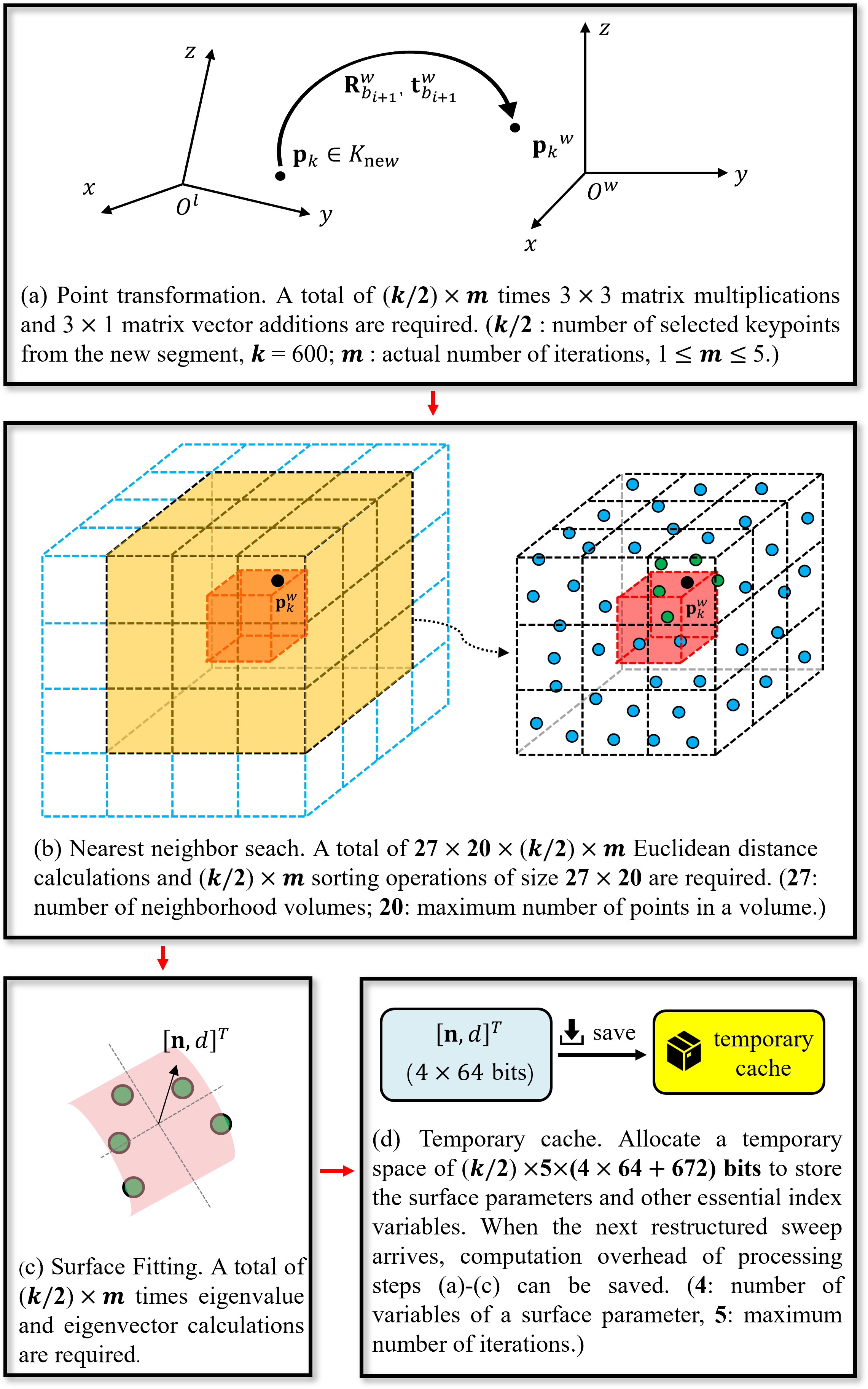}
	\end{center}
	\caption{Schematic diagram of surface parameter reuse mechanism. (a)-(d) demonstrate the sequential processing steps applied to the new segment. During subsequent state update iterations, the current new segment transitions to the old segment classification, enabling direct retrieval of its corresponding surface parameters from the temporary cache, thereby eliminating redundant computational overhead.}
	\label{fig5}
\end{figure}

The computational analysis reveals that steps 1) through 3) serve primarily to derive surface parameters for subsequent utilization in step 4). For a given 3D point, disregarding minor transformation-induced errors, repeated nearest neighbor searches yield identical results, consequently producing consistent surface parameters. Furthermore, as established in the Sweep Reconstruction section, our approach decomposes each reconstructed sweep into distinct old and new segments. Notably, the old segment has already undergone complete processing (steps 1-4) during the previous state update iteration. This characteristic enables the storage and retrieval of previously computed surface parameters, thereby eliminating redundant computations in subsequent iterations. The proposed optimization strategy, illustrated in Fig. \ref{fig5}, implements a balanced keypoint allocation scheme, assigning $k/2$ points each to the old ($K_{\text{old}}$) and new ($K_{\text{new}}$) segments. For each point $\mathbf{p}_{k_{\text{new}}}^w$ in $K_{\text{new}}$, the transformation from local to world coordinates (step 1) requires $(k/2)$$\times$5 third-order matrix operations. The subsequent nearest neighbor search (step 2) involves 27$\times$20$\times$$(k/2)$$\times$$m$ distance computations and $(k/2)$$\times$$m$ sorting operations of size 27$\times$20. Surface parameter estimation (step 3) necessitates $(k/2)$$\times$$m$ eigenvalue decompositions. The computed surface parameters for new segment points are stored in a predefined global cache, while simultaneously retrieving precomputed surface parameters for the old segment. These combined surface parameters are then utilized in step 4) for error state increment computation, completing the optimized iterative update process.

Since the keypoint selection is performed in every iteration, the $k/2$ selected points vary for the new segment, resulting in varying plane parameters. Consequently, it is necessary to record both the surface parameters and other essential variables for each iteration. The memory requirement for storing surface parameters and point indices are calculated based on their composition: each parameter comprises a 3$\times$64-bit normal vector and a 1$\times$64-bit offset, and other essential variables occupies 672 bits, resulting in a total cache size of $(k/2)$$\times$5$\times$(4$\times$64+672) bits (0.166\,MB). This implementation demonstrates significant computational efficiency compared to SR-LIO, achieving up to 50$\%$ reduction in computational overhead for steps 1) through 3) while maintaining minimal memory footprint. A comprehensive quantitative comparison of computational requirements between SR-LIO and our proposed SR-LIO++ approach is presented in Table \ref{table3}, providing detailed insights into the optimization gains.

\begin{table*}[t]
	\small\sf\centering
	\caption{Computation comparison on the first three steps of iterative state update.\label{table3}}
	\vspace{-0.2cm}
	\begin{center}
		\begin{threeparttable}
			\begin{tabular}{c|c|cc|cc}
				\toprule
				\multicolumn{1}{c|}{\multirow{2}{*}{Step}}  & \multirow{2}{*}{Calculation Type}  & \multicolumn{2}{c|}{Number of Operations} & \multicolumn{2}{c}{Additional Memory}         \\ \cline{3-6} 
				\multicolumn{1}{c|}{}                       &                                    & SR-LIO              & SR-LIO++            & SR-LIO             & SR-LIO++                 \\ \hline
				1) point transformation                     & matrix multiplication and addition & $\boldsymbol{k}$$\times$$m$               & $(\boldsymbol{k/2})$$\times$$m$               & \multirow{4}{*}{-} & \multirow{4}{*}{0.166MB} \\ \cline{1-4}
				\multirow{2}{*}{2) nearest neighbor search} & sum of squares operation           & 27$\times$20$\times$$\boldsymbol{k}$$\times$$m$         & 27$\times$20$\times$$(\boldsymbol{k/2})$$\times$$m$         &                    &                          \\ \cline{2-4}
				& sorting operation of size 27$\times$20    & $\boldsymbol{k}$$\times$$m$               & $(\boldsymbol{k/2})$$\times$$m$               &                    &                          \\ \cline{1-4}
				3) surface fitting                          & eigenvalue decomposition           & $\boldsymbol{k}$$\times$$m$               & $(\boldsymbol{k/2})$$\times$$m$               &                    &                          \\ \bottomrule
			\end{tabular}
		\end{threeparttable}
		\begin{tablenotes}
			\footnotesize
			\item[] \textbf{Denotations}: $\boldsymbol{k}$: number of keypoints per reconstructed sweep ($k=600$); $\boldsymbol{k/2}$: number of keypoint per segment; $\boldsymbol{m}$: actual number of iterations ($1 \leq m \leq 5$); \textbf{27}: number of neighborhood volumes; \textbf{20}: maximum number of points in a volume.
		\end{tablenotes}
	\end{center}
\end{table*}

While the surface parameter reutilization strategy effectively reduces computational redundancy with minimal memory overhead, it introduces a potential limitation regarding information utilization. In SR-LIO, the random selection of $k$ keypoints from each reconstructed sweep ensures uniform probability distribution of information usage. However, SR-LIO++ requires strict correspondence between the current $K_\text{old}$ and the previous $K_\text{new}$, thereby eliminating the random distribution characteristic of the old segment's information utilization. Empirical analysis reveals that the state estimation accuracy and system robustness demonstrate negligible sensitivity to this modification in information distribution. Comprehensive experimental validation of this observation is presented in the Ablation of Surface Parameter Reutilization section, providing quantitative evidence supporting the effectiveness of our approach.

Furthermore, the iterative state update process may terminate prematurely when the error state increment in the $m$-th iteration ($m < 5$) falls below a predefined threshold, regardless of whether the maximum iteration count has been reached. This early termination scenario introduces a potential limitation: during subsequent state updates, if the process proceeds to the $u$-th iteration ($u > m$), the temporary cache will lack the necessary surface parameters due to the previous early termination. To address this contingency, we randomly select $k/2$ keypoints from both the new frame and the old frame, and perform nearest neighbor search and surface fitting on all selected keypoints.

\subsubsection{Quantized Map Point Management}
\label{Quantized Map Point Management}

\begin{figure}
	\setlength{\fboxsep}{0pt}%
	\setlength{\fboxrule}{0pt}%
	\begin{center}
		\includegraphics[width=0.97\linewidth]{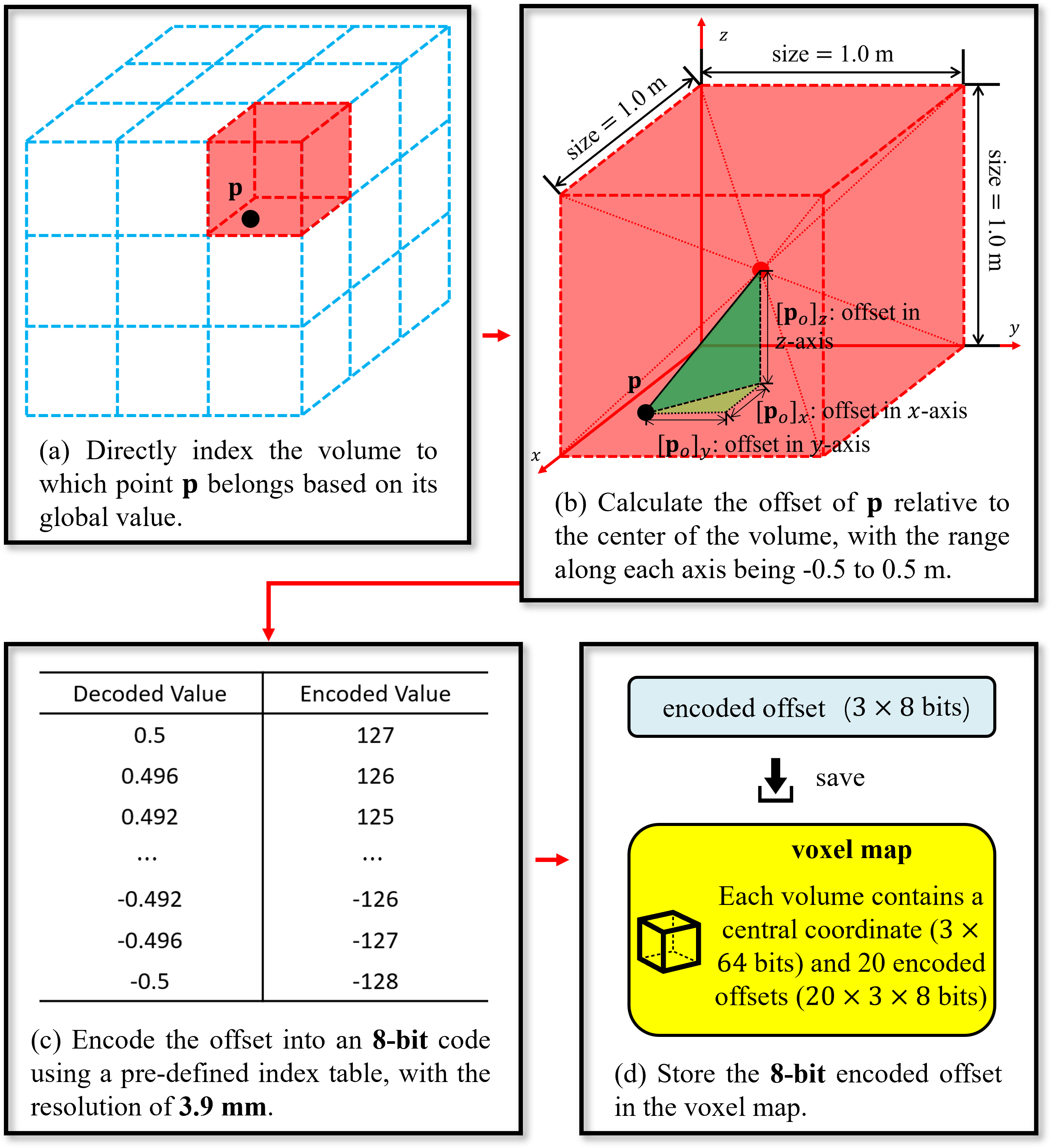}
	\end{center}
	\caption{Schematic diagram of quantized map point management utilizing index table mapping. The process initiates with the computation of spatial offsets between each global 3D point and its corresponding volume center, constrained within the range of -0.5\,m to 0.5\,m. These offsets are subsequently quantized into 8-bit char type numerical representations through a predefined index table with 3.9\,mm resolution. The quantized data is then stored in their respective volume structures. This quantization approach achieves dual optimization: 1) significant reduction in global map storage requirements, and 2) enhanced computational efficiency for nearest neighbor search operations.}
	\label{fig6}
\end{figure}

The linear growth of global map size with exploration extent poses a significant challenge for LiDAR-Inertial Odometry (LIO) systems. Conventional quantization methods can reduce point representation from 64-bit double precision to 32-bit floating point format, which is the minimum floating-point representation supported by standard CPUs. In contrast, we propose an innovative quantization approach based on index table mapping, which achieves substantial data compression by representing global map points using 8-bit char type, yielding dual benefits: 1) significant reduction in memory requirements for global map storage, and 2) improved computational efficiency in nearest neighbor search operations through reduced bit-width processing.

\begin{table*}[t]
	\small\sf\centering
	\caption{Computation comparison of distance calculation during nearest neighbor search between different management methods.\label{table4}}
	\vspace{-0.2cm}
	\begin{center}
		\begin{threeparttable}
			\begin{tabular}{p{3.2cm}<{\centering}p{4.1cm}<{\centering}|p{3.2cm}<{\centering}p{4.1cm}<{\centering}}
				\toprule
				\multicolumn{2}{c|}{Standard Map Point Management}                                                                                 & \multicolumn{2}{c}{Quantized Map Point Management}                                                                              \\ \hline
				\multicolumn{1}{c|}{Operation}   & Computation                                                                                     & \multicolumn{1}{c|}{Operation}   & Computation                                                                                  \\ \hline
				\multicolumn{1}{c|}{Calculate $\Delta x$, $\Delta y$, $\Delta z$} & \begin{tabular}[c]{@{}c@{}}27$\times$20$\times$$(k/2)$$\times$$m$$\times$3 “+” in\\ \textbf{64-bit double} domain\end{tabular}       & \multicolumn{1}{c|}{Calculate $\Delta x_e$, $\Delta y_e$, $\Delta z_e$} & \begin{tabular}[c]{@{}c@{}}27$\times$20$\times$$(k/2)$$\times$$m$$\times$3 “+” in\\ \textbf{16-bit short} domain\end{tabular}     \\ \hline
				\multicolumn{1}{c|}{Calculate $\Delta x^2$, $\Delta y^2$, $\Delta z^2$} & \begin{tabular}[c]{@{}c@{}}27$\times$20$\times$$(k/2)$$\times$$m$$\times$3 “$\times$”\\ in \textbf{64-bit double} domain\end{tabular} & \multicolumn{1}{c|}{Calculate $\Delta {x_e}^2$, $\Delta {y_e}^2$, $\Delta {z_e}^2$} & \begin{tabular}[c]{@{}c@{}}27$\times$20$\times$$(k/2)$$\times$$m$$\times$3 “$\times$”\\ in \textbf{32-bit int} domain\end{tabular} \\ \hline
				\multicolumn{1}{c|}{Calculate $\Delta d^2$} & \begin{tabular}[c]{@{}c@{}}27$\times$20$\times$$(k/2)$$\times$$m$$\times$2 “+” in\\ \textbf{64-bit double} domain\end{tabular}       & \multicolumn{1}{c|}{Calculate $\Delta {d_e}^2$} & \begin{tabular}[c]{@{}c@{}}27$\times$20$\times$$(k/2)$$\times$$m$$\times$2 “+” in\\ \textbf{32-bit int} domain\end{tabular}       \\ \bottomrule
			\end{tabular}
		\end{threeparttable}
		\begin{tablenotes}
			\footnotesize
			\item[] \textbf{Denotations}: $\boldsymbol{k/2}$: number of keypoints of a segment ($k=600$); $\boldsymbol{m}$: actual number of iterations ($1 \leq m \leq 5$); \textbf{27}: number of neighborhood volumes; \textbf{20}: number of points in a volume. The computational efficiency of numerical operations exhibits two key characteristics: 1) lower-bitwidth operations demonstrate superior execution speed compared to higher-bitwidth counterparts, and 2) for operations with equivalent bitwidth, integer arithmetic outperforms floating-point computation in terms of processing speed.
		\end{tablenotes}
	\end{center}
\end{table*}

As depicted in Fig. \ref{fig6}, the quantization process begins with volume indexing for each global 3D point $\mathbf{p}$. Leveraging the volume center coordinates stored as 3$\times$64-bit variables (as detailed in the Voxel Map Management section), we compute the spatial offset $\mathbf{p}_o=\left[\left[\mathbf{p}_o\right]_x,\left[\mathbf{p}_o\right]_y,\left[\mathbf{p}_o\right]_z\right]^T$ between $\mathbf{p}$ and its volume center. Given the fixed volume dimensions of 1.0$\times$1.0$\times$1.0 (unit: m), each component of $\mathbf{p}_o$ is constrained to the interval $[-0.5,0.5]$ meters. The core quantization process, illustrated in Fig. \ref{fig6} (c), employs a predefined index table to map these 64-bit double precision offsets to compact 8-bit char representations. This transformation is mathematically expressed as:
{\small
\begin{equation}
	\label{equation16}
	\mathbf{p}_{o e}=\operatorname{char}(\operatorname{round}\left(\frac{\mathbf{p}_o}{0.0039}\right))
\end{equation}
}
where $\operatorname{round}$ denotes integer rounding and $\operatorname{char}$ represents explicit type conversion to 8-bit char format, achieving a resolution of 3.9\,mm (0.0039\,m).

It is worth emphasizing that the construction of the index table is performed during the initialization phase. As a result, it does not incur any additional computational overhead at runtime. Once the index table is established, the encoding value can be directly indexed based on the input offset (i.e., $\operatorname{map}(\cdot)$ operation), or conversely, the offset value can be directly retrieved through reverse indexing based on the encoding value (i.e., $\operatorname{map}^{-1}(\cdot)$ operation). While this quantization introduces minor precision loss, extensive experimental validation in the Ablation of Quantized Map Point Management section confirms that the impact on state estimation accuracy remains within acceptable bounds.

\subsubsection{Nearest Neighor Search for New Segment in Short/Int Domain}
\label{Nearest Neighor Search for New Segment in Short/Int Domain}

\begin{figure}[t]
	\setlength{\fboxsep}{0pt}%
	\setlength{\fboxrule}{0pt}%
	\begin{center}
		\includegraphics[width=0.97\linewidth]{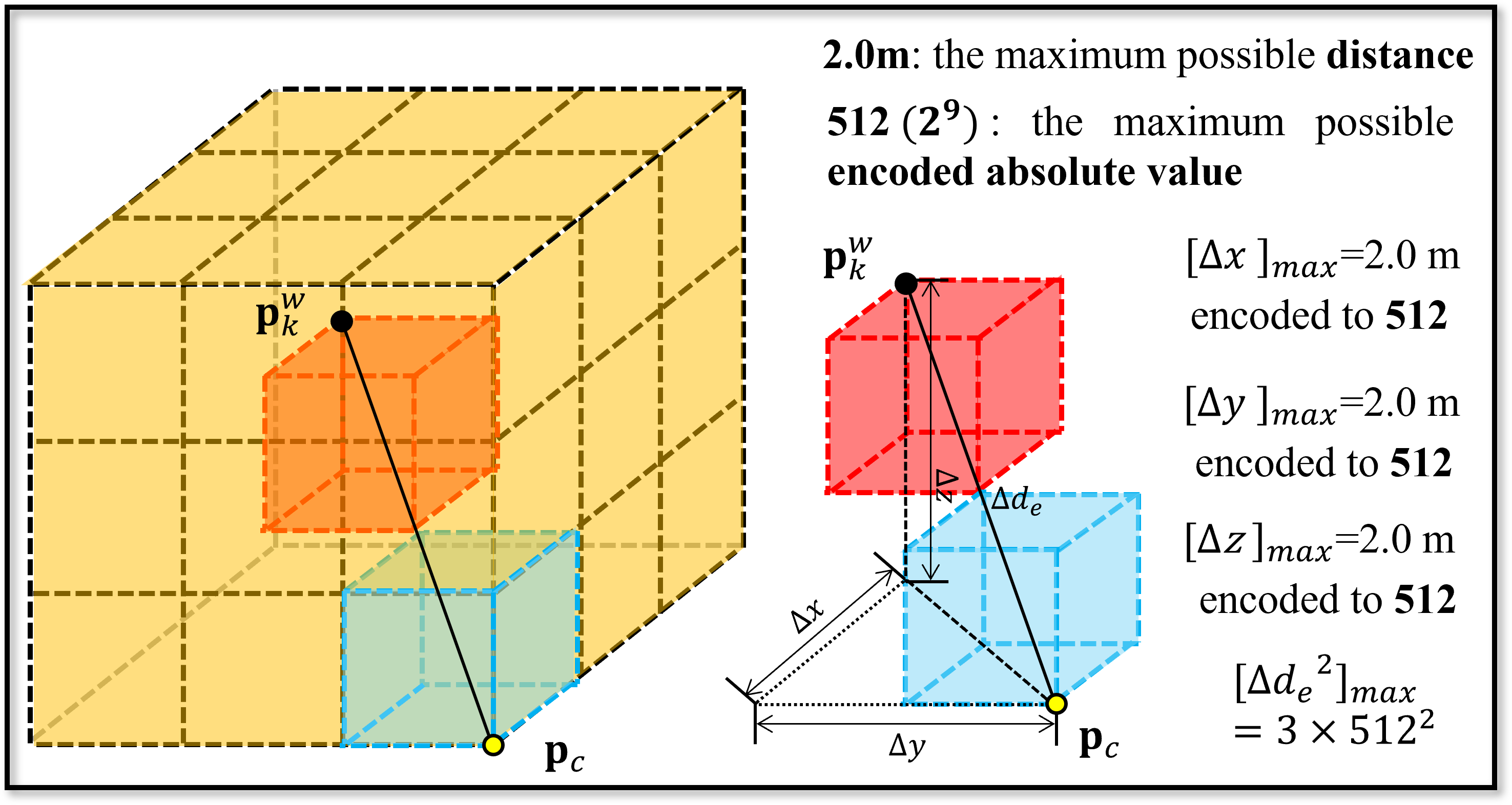}
	\end{center}
	\caption{Schematic illustration of maximum bit-width requirements for variables in 27-neighborhood nearest neighbor search using encoded offset values. The process involves computing encoded differences ($\Delta x$, $\Delta y$, $\Delta z$) within a 16-bit short integer domain, followed by the calculation of squared distance $\Delta {d_{e}}^2$ in a 32-bit integer domain. This representation demonstrates the optimal bit-width allocation for efficient nearest neighbor search operations.}
	\label{fig7}
\end{figure}

The utilization of encoded offset values in the global voxel map representation enables efficient nearest neighbor search operations. The 8-bit char type encoding of offset values significantly enhances computational efficiency for Euclidean distance calculations compared to conventional 64-bit double or 32-bit float representations, thereby accelerating the nearest neighbor search process.

We conducted a comprehensive analysis of the maximum bit-width requirements for all variables involved in the 27-neighborhood nearest neighbor search using encoded offset values. Fig. \ref{fig7} demonstrates the worst-case scenario for bit-width requirements throughout the search process. In this scenario, the maximum spatial separation occurs when the distance components ($\Delta x$, $\Delta y$, $\Delta z$) between $\mathbf{p}_k^w$ and candidate point $\mathbf{p}_c$ each measure 2.0\,m, resulting in encoded values ($\Delta x_e$, $\Delta y_e$, $\Delta z_e$) of 512 (2$^9$). This maximum value necessitates a 9-bit integer representation. To align with standard CPU architecture requirements where variable bit sizes must be multiples of 8, we employ 16-bit short type variables for storing $\Delta x_e$, $\Delta y_e$ and $\Delta z_e$, which are computed as follows:
{\small
\begin{equation}
	\label{equation17-1}
	\left[\begin{array}{l}
		\Delta x_e \\
		\Delta y_e \\
		\Delta z_e
	\end{array}\right]=\left[\mathbf{p}_k^w\right]_{o e}-(\left[\mathbf{p}_c\right]_{o e}+
	\left[\begin{array}{l}
		I_x \\
		I_y \\
		I_z
	\end{array}\right]*256)
\end{equation}
}
{\small
\begin{equation}
	\label{equation17-2}
	\left\{I_x, I_y, I_z\right\} \in\{-1,0,1\}
\end{equation}
}
where $\left[\mathbf{p}_k^w\right]_{o e}$ and $\left[\mathbf{p}_c\right]_{o e}$ represent the encoded offsets of $\mathbf{p}_k^w$ and $\mathbf{p}_c$, respectively. The values of $\left\{I_x, I_y, I_z\right\}$ are determined by the relative spatial configuration of the volumes containing $\left[\mathbf{p}_k^w\right]_{o e}$ and $\left[\mathbf{p}_c\right]_{o e}$. Eq. \ref{equation17-1} and \ref{equation17-2} demonstrates the transformation from three 64-bit double-precision subtractions to computationally efficient 16-bit short integer operations. Following this transformation, we compute the squared Euclidean distance $\Delta {d_{e}}^2$ using the encoded offsets through the following formulation:
{\small
\begin{equation}
	\label{equation18}
	\Delta {d_e}^2=\Delta {x_e}^2+\Delta {y_e}^2+\Delta {z_e}^2
\end{equation}
}
Given that $\Delta x_e$, $\Delta y_e$, and $\Delta z_e$ each equal  2$^9$, their squared values ($\Delta {x_e}^2$, $\Delta {y_e}^2$, $\Delta {z_e}^2$) consequently become 2$^{18}$, resulting in a final $\Delta {d_e}^2$ exceeding 2$^{19}$. This computational requirement necessitates type conversion of $\Delta x_e$, $\Delta y_e$ and $\Delta z_e$ from 16-bit short to 32-bit int prior to executing the three multiplications and two additions specified in Eq. \ref{equation18}. Although both int and float types utilize 32-bit representations, integer operations maintain a significant computational speed advantage over their floating-point counterparts. A comprehensive quantitative comparison between standard and quantized map point management for nearest neighbor search is presented in Table \ref{table4}.

\subsubsection{Surface Fitting for New Segment}
\label{Surface Fitting for New Segment}

\begin{figure}
	\setlength{\fboxsep}{0pt}%
	\setlength{\fboxrule}{0pt}%
	\begin{center}
		\includegraphics[width=0.97\linewidth]{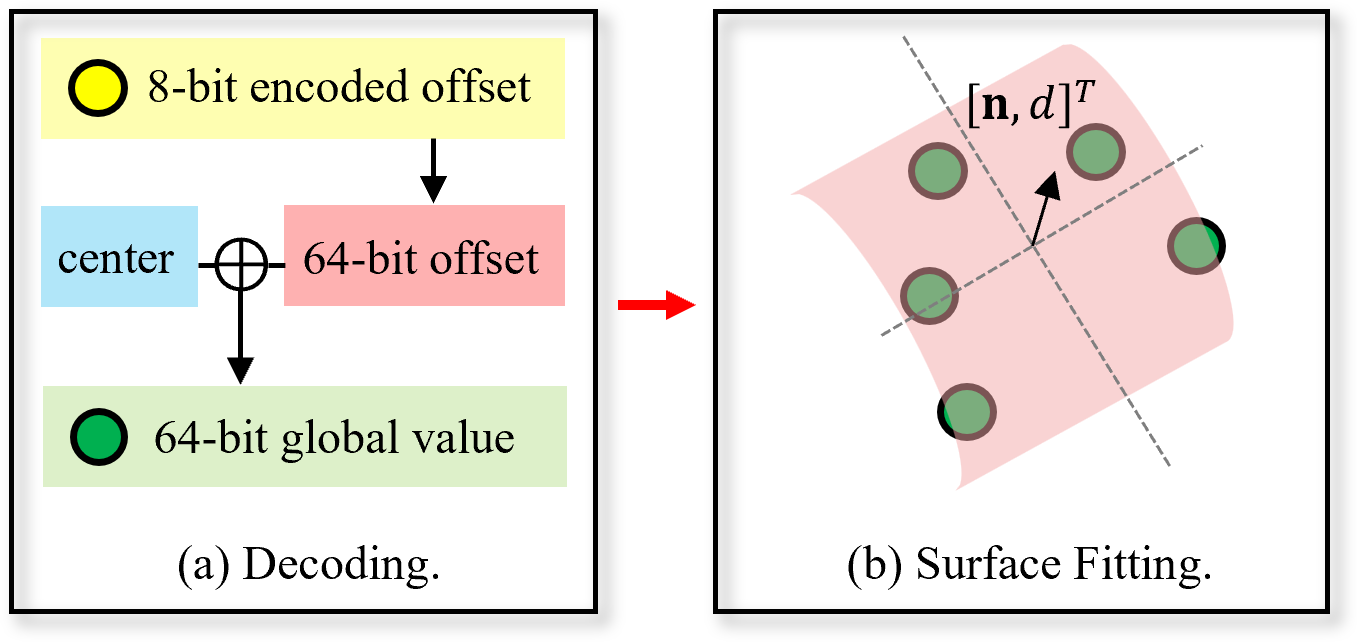}
	\end{center}
	\caption{Schematic diagram of surface fitting implementation using quantized map point management. The process initiates with the transformation of encoded offset values to global coordinates, requiring $(k/2)$$\times$20$\times$$m$ address indexing operations and 64 double-precision additions, where $k/2$ represents the number of segment keypoints, 20 corresponds to the number of nearest neighbors, and $m$ denotes the actual iteration count. This decoding step precedes the surface parameter estimation phase, ensuring accurate geometric representation.}
	\label{fig8}
\end{figure}

Following the identification of 20 nearest neighbors for each  $\mathbf{p}_k^w$ ($\mathbf{p}_k \in K_{\text{new}}$), the system proceeds with eigenvalue decomposition to derive surface parameters. As illustrated in Fig. \ref{fig8}, this computational stage necessitates conversion from encoded offset representation to global 3D coordinates. The decoding process transforms the encoded offset values back to their original global coordinate representation through the following procedure:
{\small
\begin{equation}
	\label{equation19}
	\mathbf{p}_o=\operatorname{map}^{-1}\left(\mathbf{p}_{o e}\right)
\end{equation}
}
{\small
\begin{equation}
	\label{equation20}
	\mathbf{p}^w=\mathbf{p}_o+\mathbf{v}_{\text {center}}
\end{equation}
}
where $\operatorname{map}^{-1}(\cdot)$ represents the inverse mapping operation utilizing the predefined index table for decoding, and $\mathbf{v}_{\text {center}}$ denotes the global 3D coordinates of the containing volume's centroid. As illustrated in Fig. \ref{fig8} (a), this additional decoding step represents the primary computational overhead introduced by the quantized map management approach compared to standard map point management. However, this overhead remains computationally manageable, requiring only $(k/2)$$\times$20$\times$$m$ decoding operations, where $k/2$ corresponds to the number of segment keypoints, 20 indicates the number of nearest neighbors per keypoint, and $m$ represents the actual iteration count. This additional computational cost is justified by the significant memory savings and overall system efficiency improvements achieved through quantization.

\subsubsection{Solution}
\label{Solution}

For each point $\mathbf{p}_k \in K_{\text {old }} \cup K_{\text {new }}$, we have obtained its corresponding surface parameter (i.e., normal vector $\mathbf{n}$ and offset $d$) via surface parameter reutilization or surface fitting for new segment. Accordingly, we can build the point-to-plane residual $r^{\mathbf{p}_k}$ for $\mathbf{p}_k$ as the observation constraint:
{\small
\begin{equation}
	\label{equation21-1}
	r^{\mathbf{p}_k}=\omega_{\mathbf{p}}\left(\mathbf{n}^T \mathbf{p}_k^w+d\right)
\end{equation}
}
{\small
\begin{equation}
	\label{equation21-2}
	\mathbf{p}_k^w=\mathbf{R}_{b_{i+1}}^w \mathbf{p}_k+\mathbf{t}_{b_{i+1}}^w
\end{equation}
}
where $\omega_{\mathbf{p}}$ is a weight parameter utilized in CT-ICP [\cite{dellenbach2022ct}], $\mathbf{R}_{b_{i+1}}^w$ is the rotation from $(\cdot)^{b_{i+1}}$ to $(\cdot)^w$ at $t_{i+1}$.

We define the optimal state calculated from state prediction as $\left.\boldsymbol{x}_{b_{i+1}}^w\right|_0$, and define the optimal state before current iteration as $\left.\boldsymbol{x}_{b_{i+1}}^w\right|_n$. According to the formula of state update, the incremental $\delta \boldsymbol{x}$ is calculated as:
{\small
\begin{equation}
	\label{equation24-1}
	\mathbf{K}=\left(\mathbf{H}^T \mathbf{V}^{-1} \mathbf{H}+\left(\mathbf{J}_n^0 \mathbf{\Sigma} {\mathbf{J}_n^{0}}^T\right)^{-1}\right)^{-1} \mathbf{H}^T \mathbf{V}^{-1}
\end{equation}
}
{\small
\begin{equation}
	\label{equation24-2}
	\delta \boldsymbol{x}=-\mathbf{K h}-(\mathbf{I}-\mathbf{K H}) \mathbf{J}_n^0\left(\left.\left.\boldsymbol{x}_{b_{i+1}}^w\right|_n \boxminus \boldsymbol{x}_{b_{i+1}}^w\right|_0\right)
\end{equation}
}
where the formula of observation matrix $\mathbf{h}$, observation Jacobian matrix $\mathbf{H}$, observation covariance matrix $\mathbf{V}$ and global error state Jacobian matrix $\mathbf{J}_n^0$ have been detailed in \cite{yuan2022sr}. After the incremental $\delta \boldsymbol{x}$ is calculated, we update the optimal state by:
{\small
\begin{equation}
	\label{equation27}
	\left.\boldsymbol{x}_{b_{i+1}}^w\right|_{n+1}=\left.\boldsymbol{x}_{b_{i+1}}^w\right|_n \boxplus \delta \boldsymbol{x}
\end{equation}
}
The Nearest Neighbor Search for New Segment in Short/Int Domain section$\sim$the Solution section are performed alternately until one of the following convergence conditions is met: 1) The maximum number of iterations is reached. 2) The magnitude of incremental is smaller than a threshold (e.g., 0.1\,degree for rotation and 0.01\,m for translation). After convergence, the covariance is updated as:
{\small
\begin{equation}
	\label{equation28}
	\mathbf{\Sigma}=\mathbf{J}_{n+1}^{0}(\mathbf{I}-\mathbf{K} \mathbf{H}) \mathbf{\Sigma} {\mathbf{J}_{n+1}^{0}}^T
\end{equation}
}

\subsubsection{Map Update}
\label{Map Update}

Upon completion of state estimation for the current reconstructed sweep, the system transforms the new segment's points into the world coordinate system $(\cdot)^w$ and integrates them into the voxel map. Prior to this integration, a crucial quantization step is performed, converting the global coordinates (e.g., $\mathbf{p}^w$) into encoded offset representations through the following transformation:
{\small
\begin{equation}
	\label{equation29}
	\mathbf{p}_o=\mathbf{p}^w-\mathbf{v}_{\text {center }}
\end{equation}
}
{\small
\begin{equation}
	\label{equation30}
	\mathbf{p}_{o e}=\operatorname{map}\left(\mathbf{p}_o\right)
\end{equation}
}
where $\operatorname{map}(\cdot)$ represents the encoding operation through a predefined index table, and $\mathbf{v}_{\text {center }}$ denotes the global 3D coordinates of the containing volume's centroid. This encoding process constitutes the primary computational overhead introduced by the quantized map point management approach during map updates. However, empirical results demonstrate that this additional computational cost is acceptable, and the overall computational cost for processing a single sweep remains reduced. If the number of points in a volume reaches 20, no additional points will be incorporated.

In contrast to SR-LIO's approach of removing distant map points after each state update, which is a computationally intensive operation due to the substantial overhead of voxel map traversal. To optimize system performance, we implement a periodic map maintenance routine that removes distant points at 50-second intervals, effectively balancing map management efficiency with computational resource utilization.

\section{Experiments}
\label{Experiments}

Following the evaluation methodology established in SR-LIO [\cite{yuan2022sr}], we conducted comprehensive experiments on four publicly available datasets: $nclt$ [\cite{carlevaris2016university}], $utbm$ [\cite{yaneu}], $ulhk$ [\cite{wen2020urbanloco}], and $kaist$ [\cite{jeong2019complex}]. The $nclt$ dataset, collected using an autonomous unmanned ground vehicle at the University of Michigan's North Campus, represents a large-scale, long-term dataset featuring a Velodyne HDL-32E LiDAR operating at 7.5\,Hz (130$\sim$140\,ms per 360-degree sweep) and a Microstrain MS25 IMU providing 50\,Hz measurements. Notably, the $nclt$ dataset presents unique challenges compared to the other three datasets ($utbm$, $ulhk$, and $kaist$), particularly in sequences where the Segway platform transitions from outdoor environments to long indoor corridors. These abrupt scene changes pose significant challenges for ICP point cloud registration, often causing system failures. Consequently, we exclude these challenging segments, typically occurring at sequence ends, from our evaluation. Consistent with SR-LIO's approach, we enhance the IMU data frequency to 100\,Hz through interpolation for improved state estimation.


\begin{table}[]
	\small\sf\centering
	\caption{Datasets of all sequences for evaluation.\label{table6}}
	\vspace{-0.2cm}
	\begin{center}
		\begin{tabular}{p{1.2cm}<{\centering}p{2.0cm}<{\centering}p{1.5cm}<{\centering}p{1.5cm}<{\centering}}
			\toprule
			& Name         & \begin{tabular}[c]{@{}c@{}}Duration\\ (min:sec)\end{tabular} & \begin{tabular}[c]{@{}c@{}}Distance\\ (km)\end{tabular} \\ \hline
			$nclt\_1$  & 2012-01-08   & 92:16                                                        & 6.4                                                     \\
			$nclt\_2$  & 2012-02-02   & 98:37                                                        & 6.5                                                     \\
			$nclt\_3$  & 2012-02-04   & 77:39                                                        & 5.5                                                     \\
			$nclt\_4$  & 2012-02-05   & 93:40                                                        & 6.5                                                     \\
			$nclt\_5$  & 2012-05-11   & 83:36                                                        & 6.0                                                     \\
			$nclt\_6$  & 2012-05-26   & 97:23                                                        & 6.3                                                     \\
			$nclt\_7$  & 2012-06-15   & 55:10                                                        & 4.1                                                     \\
			$nclt\_8$  & 2012-08-04   & 79:27                                                        & 5.5                                                     \\
			$nclt\_9$  & 2012-08-20   & 88:44                                                        & 6.0                                                     \\
			$nclt\_10$ & 2012-09-28   & 76:40                                                        & 5.6                                                     \\
			$nclt\_11$ & 2012-12-01   & 75:50                                                        & 5.0                                                     \\
			$utbm\_1$  & 2018-07-19   & 15:26                                                        & 4.98                                                    \\
			$utbm\_2$  & 2019-01-31   & 16:00                                                        & 6.40                                                    \\
			$utbm\_3$  & 2019-04-18   & 11:59                                                        & 5.11                                                    \\
			$utbm\_4$  & 2018-07-20   & 16:45                                                        & 4.99                                                    \\
			$utbm\_5$  & 2018-07-13   & 16:59                                                        & 5.03                                                    \\
			$ulhk\_1$  & 2019-01-17   & 5:18                                                         & 0.60                                                    \\
			$ulhk\_2$  & 2019-04-26-1 & 2:30                                                         & 0.55                                                    \\
			$kaist\_1$ & urban\_07    & 9:16                                                         & 2.55                                                    \\
			$kaist\_2$ & urban\_08    & 5.07                                                         & 1.56                                                    \\
			$kaist\_3$ & urban\_13    & 24.14                                                        & 2.36                                                    \\ \bottomrule
		\end{tabular}
	\end{center}
\end{table}

The $utbm$ dataset comprises dual 10\,Hz Velodyne HDL-32E LiDARs and a 100\,Hz Xsens MTi-28A53G25 IMU, with our implementation utilizing data from the left LiDAR. The $ulhk$ dataset provides 10\,Hz LiDAR scans from a Velodyne HDL-32E and 100\,Hz IMU measurements from a 9-axis Xsens MTi-10 IMU. The $kaist$ dataset features two 10\,Hz Velodyne VLP-16 LiDARs mounted at approximately $45^{\circ}$ tilt angles and a 200\,Hz Xsens MTi-300 IMU. Following SR-LIO's methodology, we integrate data from both LiDARs in the $kaist$ dataset. All sequences in $utbm$, $ulhk$, and $kaist$ were collected in structured urban environments using human-operated vehicles. 

\begin{table}[]
	\small\sf\centering
	\caption{RMSE of ATE comparisons (unit: m).\label{table7}}
	\vspace{-0.2cm}
	\begin{center}
		\begin{threeparttable}
			\begin{tabular}{p{0.9cm}<{\centering}|p{0.5cm}<{\centering}p{0.5cm}<{\centering}p{0.6cm}<{\centering}p{0.6cm}<{\centering}|p{0.6cm}<{\centering}p{0.6cm}<{\centering}|p{0.6cm}<{\centering}}
				\toprule
				& \begin{tabular}[c]{@{}c@{}}Fast-\\ LIO2\end{tabular} & DLIO           & \begin{tabular}[c]{@{}c@{}}Point-\\ LIO\end{tabular} & \begin{tabular}[c]{@{}c@{}}IG-\\ LIO\end{tabular} & \begin{tabular}[c]{@{}c@{}}SR-\\ LIO*\end{tabular} & \begin{tabular}[c]{@{}c@{}}SR-\\ LIO\end{tabular} & Ours           \\ \hline
				$nclt\_1$  & 3.57                                                 & 3.27           & 2.55                                                 & 1.85                                              & 1.40                                               & \textbf{1.34}                                     & \textbf{1.34}  \\
				$nclt\_2$  & 2.00                                                 & 1.80           & 2.45                                                 & 1.72                                              & 1.94                                                                                                     & 1.80                                              & \textbf{1.56}  \\
				$nclt\_3$  & 2.77                                                 & 5.35           & 5.31                                                 & 2.92                                              & 5.25                                                                                                     & 2.37                                              & \textbf{2.25}  \\
				$nclt\_4$  & 3.60                                                 & 18.10          & 1.73                                                 & 1.56                                              & $\times$                                                                                                            & 1.91                                              & \textbf{1.54}  \\
				$nclt\_5$  & 2.46                                                 & 3.14           & 11.24                                                & 1.84                                              & $\times$                                                                                                           & \textbf{1.62}                                     & 1.92           \\
				$nclt\_6$  & 2.60                                                 & 12.44          & 14.89                                                & 2.12                                              & 2.24                                                                                                      & \textbf{2.10}                                     & 2.24           \\
				$nclt\_7$  & 2.37                                                 & 2.98           & 4.39                                                 & \textbf{1.82}                                     & 2.23                                                                                                      & 2.13                                              & 1.96  \\
				$nclt\_8$  & 2.59                                                 & 7.84           & 16.28                                                & \textbf{2.40}                                     & $\times$                                                                                                          & 2.70                                              & 2.44           \\
				$nclt\_9$  & 4.01                                                 & 2.46           & 10.59                                                & \textbf{1.68}                                     & 4.00                                                                                                      & 2.11                                              & 2.35           \\
				$nclt\_10$ & 2.65                                                 & 7.72           & 16.22                                                & 1.72                                              & 1.97                                                                                                     & 1.67                                              & \textbf{1.63}  \\
				$nclt\_11$ & 4.37                                                 & 3.89           & 10.78                                                & 1.89                                              & 1.85                                                                                                      & 1.61                                              & \textbf{1.54}  \\
				$utbm\_1$  & 15.13                                                & 14.25          & 22.71                                                & 17.37                                             & 10.74                                                                                                     & 7.70                                              & \textbf{7.29}  \\
				$utbm\_2$  & 21.21                                                & \textbf{13.85} & 23.02                                                & 21.27                                             & 16.98                                                                                                      & 16.28                                             & 15.25          \\
				$utbm\_3$  & 10.81                                                & 55.28          & 13.81                                                & 13.75                                             & 9.94                                                                                                    & \textbf{8.42}                                     & 11.31          \\
				$utbm\_4$  & 15.20                                                & 18.05          & 21.76                                                & 16.44                                             & 11.35                                                                                                  & 11.12                                             & \textbf{10.46} \\
				$utbm\_5$  & 13.24                                                & 14.95          & 19.88                                                & $\times$                                                 & 10.04                                                                                                  & 9.14                                              & \textbf{8.48}  \\
				$ulhk\_1$  & 1.20                                                 & 2.44           & 1.07                                                 & 1.15                                              & 1.02                                                                                                     & \textbf{0.93}                                     & 1.13           \\
				$ulhk\_2$  & 3.24                                                 & $\times$              & \textbf{2.82}                                        & 3.31                                              & 3.36                                                                                                     & 3.21                                              & 3.18           \\
				$kaist\_1$ & 0.88                                                 & 1.04  & \textbf{0.75}                                                 & 61.20                                             & 1.14                                                                                                    & 1.10                                              & 1.12           \\
				$kaist\_2$ & 16.27                                                & 1.91           & 1.08                                                 & 3.01                                              & 0.95                                                                                                    & 0.92                                              & \textbf{0.87}  \\
				$kaist\_3$ & $\times$                                                    & $\times$              & 3.04                                                 & $\times$                                                 & 1.56                                                                                                  & \textbf{1.36}                                     & \textbf{1.36}  \\ \bottomrule
			\end{tabular}
		\end{threeparttable}
		\begin{tablenotes}
			\footnotesize
			\item[] \textbf{Denotations}: “$\times$” means the system fails to run entirety on the corresponding sequence. “SR-LIO*” denotes a variant of the SR-LIO system framework that excludes sweep reconstruction, thus maintaining the original output frequency of 10\,Hz.
		\end{tablenotes}
	\end{center}
\end{table}

Our evaluation employs the same 21 test sequences utilized in SR-LIO, with detailed sequence information provided in Table \ref{table6}. For quantitative accuracy assessment, we adopt the Absolute Translational Error (ATE) metric. As SR-LIO++ is fundamentally a single-threaded system, its accuracy exhibits no essential variation across different computing platforms. Thus, all ATE evaluations were conducted on a PC equipped with an Intel Core i9-14900HX processor. To demonstrate the practical applicability of our approach across different robotic platforms, we extended our evaluation to two additional computing environments: 1) an industrial control computer with an Intel Xeon Platinum 8352V CPU, representing typical hardware for composite robots, and 2) a Raspberry Pi 4B. The Raspberry Pi 4B evaluation specifically demonstrates our method's capability to maintain a 20\,Hz output frequency on resource-constrained platforms. Our method requires no adjustment of hyperparameters when tested across different sequences.

\subsection{ATE Comparison of the State-of-the-Arts}
\label{ATE Comparison of the State-of-the-Arts}

\begin{table*}[]
	\small\sf\centering
	\caption{Time comparison with baseline (unit: ms).\label{table8}}
	\vspace{-0.2cm}
	\begin{center}
		\begin{threeparttable}
			\begin{tabular}{c|p{1.1cm}<{\centering}p{0.8cm}<{\centering}p{0.8cm}<{\centering}p{0.8cm}<{\centering}|p{1.1cm}<{\centering}p{0.8cm}<{\centering}p{0.8cm}<{\centering}p{0.8cm}<{\centering}|p{1.1cm}<{\centering}p{0.8cm}<{\centering}p{0.8cm}<{\centering}p{0.8cm}<{\centering}}
				\toprule
				Platform & \multicolumn{4}{c|}{Intel-Core i9-14900HX} & \multicolumn{4}{c|}{Intel Xeon Platinum 8352V} & \multicolumn{4}{c}{Raspberry Pi 4B}     \\ \hline
				Method   & SR-LIO  & Base   & Base+  & Ours           & SR-LIO   & Base    & Base+   & Ours            & SR-LIO & Base  & Base+ & Ours           \\ \hline
				$nclt\_1$  & \cellcolor{green!20}27.2   & \cellcolor{green!20}17.9  & \cellcolor{green!20}17.0  & \cellcolor{green!20}\textbf{15.1} & \cellcolor{green!20}40.0    & \cellcolor{green!20}26.5   & \cellcolor{green!20}24.2   & \cellcolor{green!20}\textbf{20.5}  & \cellcolor{red!20}79.6  & \cellcolor{red!20}58.6 & \cellcolor{red!20}50.4 & \cellcolor{green!20}\textbf{45.1} \\
				$nclt\_2$  & \cellcolor{green!20}25.3   & \cellcolor{green!20}17.9  & \cellcolor{green!20}16.8  & \cellcolor{green!20}\textbf{14.9} & \cellcolor{green!20}36.3    & \cellcolor{green!20}26.3   & \cellcolor{green!20}23.8   & \cellcolor{green!20}\textbf{19.9}  & \cellcolor{red!20}69.6  & \cellcolor{red!20}51.9 & \cellcolor{green!20}45.9 & \cellcolor{green!20}\textbf{40.9} \\
				$nclt\_3$  & \cellcolor{green!20}24.5   & \cellcolor{green!20}17.9  & \cellcolor{green!20}17.2  & \cellcolor{green!20}\textbf{15.6} & \cellcolor{green!20}35.8    & \cellcolor{green!20}26.2   & \cellcolor{green!20}24.3   & \cellcolor{green!20}\textbf{20.9}  & \cellcolor{red!20}78.7  & \cellcolor{red!20}55.8 & \cellcolor{green!20}48.0 & \cellcolor{green!20}\textbf{42.3} \\
				$nclt\_4$  & \cellcolor{green!20}26.0   & \cellcolor{green!20}18.5  & \cellcolor{green!20}17.6  & \cellcolor{green!20}\textbf{15.6} & \cellcolor{green!20}40.1    & \cellcolor{green!20}27.1   & \cellcolor{green!20}24.7   & \cellcolor{green!20}\textbf{21.0}  & \cellcolor{red!20}80.3  & \cellcolor{red!20}58.3 & \cellcolor{red!20}51.2 & \cellcolor{green!20}\textbf{45.8} \\
				$nclt\_5$  & \cellcolor{green!20}26.0   & \cellcolor{green!20}18.6  & \cellcolor{green!20}18.5  & \cellcolor{green!20}\textbf{16.4} & \cellcolor{green!20}40.2    & \cellcolor{green!20}28.3   & \cellcolor{green!20}26.5   & \cellcolor{green!20}\textbf{22.2}  & \cellcolor{red!20}89.8  & \cellcolor{red!20}64.9 & \cellcolor{red!20}54.8 & \cellcolor{green!20}\textbf{48.5} \\
				$nclt\_6$  & \cellcolor{green!20}25.7   & \cellcolor{green!20}19.2  & \cellcolor{green!20}17.9  & \cellcolor{green!20}\textbf{16.0} & \cellcolor{green!20}39.3    & \cellcolor{green!20}28.1   & \cellcolor{green!20}25.4   & \cellcolor{green!20}\textbf{21.2}  & \cellcolor{red!20}83.2  & \cellcolor{red!20}59.8 & \cellcolor{red!20}50.7 & \cellcolor{green!20}\textbf{45.1} \\
				$nclt\_7$  & \cellcolor{green!20}24.2   & \cellcolor{green!20}18.0  & \cellcolor{green!20}16.9  & \cellcolor{green!20}\textbf{15.2} & \cellcolor{green!20}35.5    & \cellcolor{green!20}27.0   & \cellcolor{green!20}24.4   & \cellcolor{green!20}\textbf{20.2}  & \cellcolor{red!20}79.5  & \cellcolor{red!20}58.1 & \cellcolor{green!20}47.7 & \cellcolor{green!20}\textbf{42.2} \\
				$nclt\_8$  & \cellcolor{green!20}27.5   & \cellcolor{green!20}20.0  & \cellcolor{green!20}19.5  & \cellcolor{green!20}\textbf{17.7} & \cellcolor{green!20}40.7    & \cellcolor{green!20}31.5   & \cellcolor{green!20}29.4   & \cellcolor{green!20}\textbf{25.1}  & \cellcolor{red!20}82.0  & \cellcolor{red!20}61.2 & \cellcolor{red!20}52.1 & \cellcolor{green!20}\textbf{47.3} \\
				$nclt\_9$  & \cellcolor{green!20}26.5   & \cellcolor{green!20}18.5  & \cellcolor{green!20}17.6  & \cellcolor{green!20}\textbf{15.9} & \cellcolor{green!20}39.7    & \cellcolor{green!20}27.2   & \cellcolor{green!20}25.2   & \cellcolor{green!20}\textbf{21.0}  & \cellcolor{red!20}85.7  & \cellcolor{red!20}60.7 & \cellcolor{red!20}52.2 & \cellcolor{green!20}\textbf{46.2} \\
				$nclt\_10$ & \cellcolor{green!20}26.1   & \cellcolor{green!20}17.9  & \cellcolor{green!20}17.2  & \cellcolor{green!20}\textbf{15.4} & \cellcolor{green!20}38.4    & \cellcolor{green!20}26.7   & \cellcolor{green!20}24.7   & \cellcolor{green!20}\textbf{20.1}  & \cellcolor{red!20}79.4  & \cellcolor{red!20}57.6 & \cellcolor{green!20}49.4 & \cellcolor{green!20}\textbf{44.0} \\
				$nclt\_11$ & \cellcolor{green!20}24.6   & \cellcolor{green!20}16.6  & \cellcolor{green!20}15.6  & \cellcolor{green!20}\textbf{13.9} & \cellcolor{green!20}34.1    & \cellcolor{green!20}24.6   & \cellcolor{green!20}21.9   & \cellcolor{green!20}\textbf{18.2}  & \cellcolor{red!20}61.5  & \cellcolor{green!20}47.4 & \cellcolor{green!20}41.5 & \cellcolor{green!20}\textbf{36.7} \\
				$utbm\_1$  & \cellcolor{green!20}20.7   & \cellcolor{green!20}15.3  & \cellcolor{green!20}13.6  & \cellcolor{green!20}\textbf{12.7} & \cellcolor{green!20}27.9    & \cellcolor{green!20}22.2   & \cellcolor{green!20}20.0   & \cellcolor{green!20}\textbf{18.0}  & \cellcolor{red!20}95.5  & \cellcolor{red!20}57.3 & \cellcolor{green!20}49.0 & \cellcolor{green!20}\textbf{43.9} \\
				$utbm\_2$  & \cellcolor{green!20}21.1   & \cellcolor{green!20}16.6  & \cellcolor{green!20}15.0  & \cellcolor{green!20}\textbf{13.9} & \cellcolor{green!20}28.4    & \cellcolor{green!20}22.6   & \cellcolor{green!20}20.6   & \cellcolor{green!20}\textbf{18.6}  & \cellcolor{red!20}92.7  & \cellcolor{red!20}59.4 & \cellcolor{green!20}49.4 & \cellcolor{green!20}\textbf{43.5} \\
				$utbm\_3$  & \cellcolor{green!20}20.5   & \cellcolor{green!20}15.9  & \cellcolor{green!20}14.2  & \cellcolor{green!20}\textbf{13.4} & \cellcolor{green!20}26.7    & \cellcolor{green!20}22.0   & \cellcolor{green!20}19.9   & \cellcolor{green!20}\textbf{17.9}  & \cellcolor{red!20}83.8  & \cellcolor{red!20}57.9 & \cellcolor{green!20}48.9 & \cellcolor{green!20}\textbf{42.3} \\
				$utbm\_4$  & \cellcolor{green!20}21.2   & \cellcolor{green!20}15.3  & \cellcolor{green!20}13.8  & \cellcolor{green!20}\textbf{12.8} & \cellcolor{green!20}28.9    & \cellcolor{green!20}22.2   & \cellcolor{green!20}19.9   & \cellcolor{green!20}\textbf{18.0}  & \cellcolor{red!20}97.7  & \cellcolor{red!20}60.4 & \cellcolor{green!20}48.9 & \cellcolor{green!20}\textbf{43.4} \\
				$utbm\_5$  & \cellcolor{green!20}20.4   & \cellcolor{green!20}14.6  & \cellcolor{green!20}13.1  & \cellcolor{green!20}\textbf{12.0} & \cellcolor{green!20}29.5    & \cellcolor{green!20}21.9   & \cellcolor{green!20}19.6   & \cellcolor{green!20}\textbf{17.7}  & \cellcolor{red!20}91.4  & \cellcolor{red!20}56.7 & \cellcolor{green!20}45.6 & \cellcolor{green!20}\textbf{41.4} \\
				$ulhk\_1$  & \cellcolor{green!20}20.8   & \cellcolor{green!20}14.6  & \cellcolor{green!20}13.0  & \cellcolor{green!20}\textbf{12.5} & \cellcolor{green!20}25.0    & \cellcolor{green!20}23.4   & \cellcolor{green!20}21.3   & \cellcolor{green!20}\textbf{20.1}  & \cellcolor{red!20}74.5  & \cellcolor{red!20}57.2 & \cellcolor{red!20}50.1 & \cellcolor{green!20}\textbf{45.5} \\
				$ulhk\_2$  & \cellcolor{green!20}20.6   & \cellcolor{green!20}14.2  & \cellcolor{green!20}12.8  & \cellcolor{green!20}\textbf{12.4} & \cellcolor{green!20}25.0    & \cellcolor{green!20}23.2   & \cellcolor{green!20}21.1   & \cellcolor{green!20}\textbf{20.3}  & \cellcolor{red!20}72.8  & \cellcolor{red!20}58.9 & \cellcolor{red!20}51.6 & \cellcolor{green!20}\textbf{46.1} \\
				$kaist\_1$ & \cellcolor{green!20}20.2   & \cellcolor{green!20}14.9  & \cellcolor{green!20}12.9  & \cellcolor{green!20}\textbf{11.8} & \cellcolor{green!20}25.9    & \cellcolor{green!20}18.3   & \cellcolor{green!20}16.2   & \cellcolor{green!20}\textbf{14.9}  & \cellcolor{red!20}83.9  & \cellcolor{green!20}45.9 & \cellcolor{green!20}36.5 & \cellcolor{green!20}\textbf{32.8} \\
				$kaist\_2$ & \cellcolor{green!20}19.7   & \cellcolor{green!20}14.8  & \cellcolor{green!20}13.6  & \cellcolor{green!20}\textbf{11.7} & \cellcolor{green!20}23.3    & \cellcolor{green!20}19.7   & \cellcolor{green!20}18.0   & \cellcolor{green!20}\textbf{15.8}  & \cellcolor{red!20}82.7  & \cellcolor{green!20}43.9 & \cellcolor{green!20}37.9 & \cellcolor{green!20}\textbf{33.1} \\
				$kaist\_3$ & \cellcolor{green!20}19.7   & \cellcolor{green!20}13.3  & \cellcolor{green!20}11.1  & \cellcolor{green!20}\textbf{9.6}  & \cellcolor{green!20}24.5    & \cellcolor{green!20}16.5   & \cellcolor{green!20}14.3   & \cellcolor{green!20}\textbf{12.1}  & \cellcolor{red!20}69.4  & \cellcolor{green!20}35.9 & \cellcolor{green!20}25.8 & \cellcolor{green!20}\textbf{23.5} \\ \bottomrule
			\end{tabular}
		\end{threeparttable}
		\begin{tablenotes}
			\footnotesize
			\item[] \textbf{Denotations}: “Base” represents the original SR-LIO with strategic optimizations, particularly the modification of the far-point removal mechanism from its original implementation in every reconstructed sweep to an optimized interval of 50 seconds. “Base+” denotes the subsequent integration of our proposed surface parameter reutilization strategy into the optimized “Base” system. We color the processing time of a single sweep as \colorbox{red!20}{non-real-time} when it exceeds 50\,ms, and as \colorbox{green!20}{real-time} when it is below 50\,ms.
		\end{tablenotes}
	\end{center}
\end{table*}

The field of LiDAR-Inertial Odometry (LIO) has witnessed significant advancements in recent years, accompanied by the proliferation of numerous open-source implementations. For our comparative analysis, we have selected five state-of-the-art LIO systems that represent the most significant developments in the field: Fast-LIO2 [\cite{xu2022fast}], DLIO [\cite{chen2023direct}], Point-LIO [\cite{he2023point}], IG-LIO [\cite{chen2024ig}] and SR-LIO [\cite{yuan2022sr}]. The comparative ATE results for these systems are obtained from the recorded data in SR-LIO [\cite{yuan2022sr}].

The experimental results presented in Table \ref{table7} demonstrate that our proposed approach achieves comparable or superior performance compared to state-of-the-art methods, as evidenced by smaller RMSE of ATE values in more than half of the test sequences. Through comprehensive evaluation, our method exhibits comparable accuracy to two recent advanced works, IG-LIO and SR-LIO. Notably, since SR-LIO serves as our baseline framework and demonstrates comparable ATE performance with our method, these experimental results substantiate that our innovative surface parameter reutilization strategy and quantized map point management mechanism maintain system accuracy and robustness without significant degradation.

Among the evaluated methods, SR-LIO and SR-LIO++ are distinguished by their ability to output optimized states at a frequency of \textbf{20\,Hz}, whereas other methods are limited to a state output frequency of \textbf{10\,Hz} during evaluation. Although Point-LIO is theoretically capable of outputting the pose at the frequency of individual point acquisition, empirical testing reveals that its most stable state output is at 10\,Hz. The reason is that the 10\,Hz frequency corresponds to the completion of a full 360-degree LiDAR scan, ensuring uniform spatial distribution and balanced utilization of the acquired data points.

\subsection{Time Consumption Comparison with Baseline}
\label{Time Consumption Comparison with Baseline}

\begin{table*}[]
	\small\sf\centering
	\caption{Ablation study of surface parameter reutilization on time consumption (unit: ms).\label{table9}}
	\vspace{-0.2cm}
	\begin{center}
		\begin{threeparttable}
			\begin{tabular}{c|p{1.0cm}<{\centering}p{1.0cm}<{\centering}p{0.75cm}<{\centering}p{0.75cm}<{\centering}|p{1.0cm}<{\centering}p{1.0cm}<{\centering}p{0.75cm}<{\centering}p{0.75cm}<{\centering}|p{1.0cm}<{\centering}p{1.0cm}<{\centering}p{0.75cm}<{\centering}p{0.75cm}<{\centering}}
				\toprule
				Platform & \multicolumn{4}{c|}{Intel-Core i9-14900HX}                             & \multicolumn{4}{c|}{Intel Xeon Platinum 8352V}                         & \multicolumn{4}{c}{Raspberry Pi 4B}                                     \\ \hline
				Module   & \multicolumn{2}{c|}{Residual Building}    & \multicolumn{2}{c|}{Total} & \multicolumn{2}{c|}{Residual Building}    & \multicolumn{2}{c|}{Total} & \multicolumn{2}{c|}{Residual Building}      & \multicolumn{2}{c}{Total} \\ \hline
				Method   & Base & \multicolumn{1}{c|}{Base+}         & Base    & Base+            & Base & \multicolumn{1}{c|}{Base+}         & Base    & Base+            & Base  & \multicolumn{1}{c|}{Base+}          & Base    & Base+           \\ \hline
				$nclt\_1$  & 6.5 & \multicolumn{1}{c|}{\textbf{5.2}} & 17.9   & \textbf{17.0}   & 8.4 & \multicolumn{1}{c|}{\textbf{6.5}} & 26.5   & \textbf{24.2}   & 16.1 & \multicolumn{1}{c|}{\textbf{12.6}} & 58.6   & \textbf{50.4}  \\
				$nclt\_2$  & 6.7 & \multicolumn{1}{c|}{\textbf{5.2}} & 17.9   & \textbf{16.8}   & 8.6 & \multicolumn{1}{c|}{\textbf{6.6}} & 26.3   & \textbf{23.8}   & 15.8 & \multicolumn{1}{c|}{\textbf{12.3}} & 51.9   & \textbf{45.9}  \\
				$nclt\_3$  & 6.4 & \multicolumn{1}{c|}{\textbf{5.2}} & 17.9   & \textbf{17.2}   & 8.4 & \multicolumn{1}{c|}{\textbf{6.5}} & 26.2   & \textbf{24.3}   & 16.1 & \multicolumn{1}{c|}{\textbf{12.6}} & 55.8   & \textbf{48.0}  \\
				$nclt\_4$  & 6.5 & \multicolumn{1}{c|}{\textbf{5.3}} & 18.5   & \textbf{17.6}   & 8.6 & \multicolumn{1}{c|}{\textbf{6.6}} & 27.1   & \textbf{24.7}   & 16.4 & \multicolumn{1}{c|}{\textbf{12.9}} & 58.3   & \textbf{51.2}  \\
				$nclt\_5$  & 5.9 & \multicolumn{1}{c|}{\textbf{5.2}} & 18.6   & \textbf{18.5}   & 8.4 & \multicolumn{1}{c|}{\textbf{6.8}} & 28.3   & \textbf{26.5}   & 16.6 & \multicolumn{1}{c|}{\textbf{13.5}} & 64.9   & \textbf{54.8}  \\
				$nclt\_6$  & 6.6 & \multicolumn{1}{c|}{\textbf{5.2}} & 19.2   & \textbf{17.9}   & 8.6 & \multicolumn{1}{c|}{\textbf{6.7}} & 28.1   & \textbf{25.4}   & 16.4 & \multicolumn{1}{c|}{\textbf{12.8}} & 59.8   & \textbf{50.7}  \\
				$nclt\_7$  & 6.2 & \multicolumn{1}{c|}{\textbf{5.1}} & 18.0   & \textbf{16.9}   & 8.6 & \multicolumn{1}{c|}{\textbf{6.6}} & 27.0   & \textbf{24.4}   & 18.0 & \multicolumn{1}{c|}{\textbf{15.4}} & 58.1   & \textbf{47.7}  \\
				$nclt\_8$  & 5.6 & \multicolumn{1}{c|}{\textbf{4.9}} & 20.0   & \textbf{19.5}   & 8.8 & \multicolumn{1}{c|}{\textbf{7.2}} & 31.5   & \textbf{29.4}   & 16.2 & \multicolumn{1}{c|}{\textbf{13.2}} & 61.2   & \textbf{52.1}  \\
				$nclt\_9$  & 6.1 & \multicolumn{1}{c|}{\textbf{5.1}} & 18.5   & \textbf{17.6}   & 8.3 & \multicolumn{1}{c|}{\textbf{6.6}} & 27.2   & \textbf{25.2}   & 16.6 & \multicolumn{1}{c|}{\textbf{12.8}} & 60.7   & \textbf{52.2}  \\
				$nclt\_10$ & 6.3 & \multicolumn{1}{c|}{\textbf{5.1}} & 17.9   & \textbf{17.2}   & 8.5 & \multicolumn{1}{c|}{\textbf{6.8}} & 26.7   & \textbf{24.7}   & 16.7 & \multicolumn{1}{c|}{\textbf{12.9}} & 57.6   & \textbf{49.4}  \\
				$nclt\_11$ & 6.3 & \multicolumn{1}{c|}{\textbf{5.0}} & 16.6   & \textbf{15.6}   & 8.2 & \multicolumn{1}{c|}{\textbf{6.1}} & 24.6   & \textbf{21.9}   & 14.7 & \multicolumn{1}{c|}{\textbf{11.1}} & 47.4   & \textbf{41.5}  \\
				$utbm\_1$  & 6.0 & \multicolumn{1}{c|}{\textbf{4.3}} & 15.3   & \textbf{13.6}   & 7.9 & \multicolumn{1}{c|}{\textbf{5.5}} & 22.2   & \textbf{20.0}   & 14.8 & \multicolumn{1}{c|}{\textbf{10.2}} & 57.3   & \textbf{49.0}  \\
				$utbm\_2$  & 5.9 & \multicolumn{1}{c|}{\textbf{4.7}} & 16.6   & \textbf{15.0}   & 7.8 & \multicolumn{1}{c|}{\textbf{5.7}} & 22.6   & \textbf{20.6}   & 14.6 & \multicolumn{1}{c|}{\textbf{10.5}} & 59.4   & \textbf{49.4}  \\
				$utbm\_3$  & 6.0 & \multicolumn{1}{c|}{\textbf{4.6}} & 15.9   & \textbf{14.2}   & 7.9 & \multicolumn{1}{c|}{\textbf{5.6}} & 22.0   & \textbf{19.9}   & 14.3 & \multicolumn{1}{c|}{\textbf{10.3}} & 57.9   & \textbf{48.9}  \\
				$utbm\_4$  & 6.1 & \multicolumn{1}{c|}{\textbf{4.4}} & 15.3   & \textbf{13.8}   & 8.0 & \multicolumn{1}{c|}{\textbf{5.5}} & 22.2   & \textbf{19.9}   & 15.2 & \multicolumn{1}{c|}{\textbf{10.2}} & 60.4   & \textbf{48.9}  \\
				$utbm\_5$  & 5.5 & \multicolumn{1}{c|}{\textbf{4.1}} & 14.6   & \textbf{13.1}   & 7.9 & \multicolumn{1}{c|}{\textbf{5.5}} & 21.9   & \textbf{19.6}   & 14.6 & \multicolumn{1}{c|}{\textbf{10.0}} & 56.7   & \textbf{45.5}  \\
				$ulhk\_1$  & 4.5 & \multicolumn{1}{c|}{\textbf{3.4}} & 14.6   & \textbf{13.0}   & 6.5 & \multicolumn{1}{c|}{\textbf{4.8}} & 23.4   & \textbf{21.3}   & 13.7 & \multicolumn{1}{c|}{\textbf{9.8}}  & 57.2   & \textbf{50.1}  \\
				$ulhk\_2$  & 4.7 & \multicolumn{1}{c|}{\textbf{3.4}} & 14.2   & \textbf{12.8}   & 6.3 & \multicolumn{1}{c|}{\textbf{5.5}} & 23.2   & \textbf{21.1}   & 13.6 & \multicolumn{1}{c|}{\textbf{10.0}} & 58.9   & \textbf{51.6}  \\
				$kaist\_1$ & 7.5 & \multicolumn{1}{c|}{\textbf{5.2}} & 14.9   & \textbf{12.9}   & 8.2 & \multicolumn{1}{c|}{\textbf{6.0}} & 18.3   & \textbf{16.2}   & 16.0 & \multicolumn{1}{c|}{\textbf{11.1}} & 45.9   & \textbf{36.5}  \\
				$kaist\_2$ & 6.5 & \multicolumn{1}{c|}{\textbf{5.1}} & 14.8   & \textbf{13.6}   & 7.5 & \multicolumn{1}{c|}{\textbf{5.9}} & 19.7   & \textbf{18.0}   & 14.3 & \multicolumn{1}{c|}{\textbf{10.6}} & 43.9   & \textbf{37.9}  \\
				$kaist\_3$ & 7.9 & \multicolumn{1}{c|}{\textbf{5.2}} & 13.3   & \textbf{11.1}   & 8.9 & \multicolumn{1}{c|}{\textbf{6.5}} & 16.5   & \textbf{14.3}   & 14.8 & \multicolumn{1}{c|}{\textbf{11.1}} & 35.9   & \textbf{25.8}  \\ \bottomrule
			\end{tabular}
		\end{threeparttable}
		\begin{tablenotes}
			\footnotesize
			\item[] \textbf{Denotations}: “Base” represents the original SR-LIO with strategic optimizations, particularly the modification of the far-point removal mechanism from its original implementation in every reconstructed sweep to an optimized interval of 50 seconds. “Base+” denotes the subsequent integration of our proposed surface parameter reutilization strategy into the optimized “Base” system. “Residual Building” refers to the computational time for constructing point-to-plane residuals in each state update iteration, comprising nearest neighbor search, surface fitting, and residual Jacobian computation. It is executed $m$ times per reconstructed sweep (where $1 \le m \le 5$). “Total” indicates the complete processing time per reconstructed sweep.
		\end{tablenotes}
	\end{center}
\end{table*}

We conduct a comprehensive evaluation of the computational overhead between our system and the baseline across three distinct hardware platforms. While several existing methods, including Fast-LIO2 and IG-LIO, demonstrate faster operational speeds than SR-LIO, none of them possess the capability to enhance the output frequency. The primary objective of our work is to enable SR-LIO, with its frequency-enhancing functionality, to achieve 20\,Hz state output on hardware platforms with limited computational resources. To this end, we have implemented a series of strategic optimizations to SR-LIO, including modifying the far-point deletion strategy from its original per-reconstructed-sweep execution to a more efficient 50-second interval implementation. This optimized system is designated as “Base”. Subsequently, we incrementally integrate our proposed surface parameter reutilization method into “Base”, resulting in an enhanced system denoted as “Base+”. Finally, we incorporate our novel quantized map point management method into “Base+”, culminating in our ultimate system “Ours”.

Results in Table \ref{table8} demonstrate the computational overhead comparison among our SR-LIO++, SR-LIO, “Base”, and “Base+”. Although the strategic optimizations yield the most significant improvement in computational efficiency, they still cannot guarantee real-time performance across all test sequences on the Raspberry Pi 4B platform. The incorporation of our proposed surface parameter reutilization method into the “Base” system further enhances computational efficiency. This improvement stems from the fact that the sweep reconstruction results in overlapping data segments between adjacent reconstructed sweeps, and “Base” indiscriminately processes the data in these overlapping segments, leading to redundant computations, “Base+” effectively avoids such unnecessary repetitive calculations. Furthermore, the addition of our quantized map point management method to "Base+" achieves sufficient computational efficiency to enable stable 20\,Hz state output and map update on the Raspberry Pi 4B platform. Compared to “Base+”, our quantized map point management method requires fewer bits to represent map points, thereby significantly improving computational efficiency during the high-frequency Euclidean distance calculations in the nearest neighbor search step.

\subsection{Ablation of Surface Parameter Reutilization}
\label{Ablation Study of Surface Parameter Reutilization}

In this section, we conduct an ablation study to systematically evaluate the impact of the surface parameter reutilization method on both computational overhead and accuracy of state estimation. As detailed in the Surface Parameter Reutilization section, this method is theoretically capable of reducing the computational load of nearest neighbor search and surface fitting operations by up to 50$\%$. In the experimental evaluation, while it would be ideal to separately measure the average time consumption of these two submodules (nearest neighbor search and surface fitting), practical constraints necessitate an alternative approach. During each iterative state update, these operations are executed $(k/2)$$\times$$m$ times (where $k/2$ represents the number of keypoints in a new segment and $m$ denotes the actual number of iterations). The high-frequency execution of these operations renders the inherent computational cost of the timing function non-negligible, potentially introducing measurement artifacts. To mitigate the potential inflation of timing measurements caused by the high-frequency execution of timing functions, we instead evaluate the time overhead of the parent function encompassing both operations, namely the “Residual Building”. This module is executed $m$ times per reconstructed sweep (where $1 \le m \le 5$), with each execution comprising: 1) nearest neighbor search and surface parameter fitting for all keypoints in the new segment, and 2) Jacobian computation of each point-to-plane residual from both old and new segments. The “Residual Building” module encapsulates all computational components where surface parameter reutilization can potentially contribute to efficiency improvements, making it an appropriate metric for assessing the overall impact of our contribution.

Experimental results presented in Table \ref{table9} confirm that our surface parameter reutilization method effectively improves the computational efficiency of the “Residual Building” module, although the observed enhancement is below the theoretical 50$\%$ reduction. This performance gap can be attributed to two primary factors: First, as discussed in the Surface Parameter Reutilization section, when the current state update requires more iterations than the previous one, the surface parameter reutilization cannot optimize the computational load for these additional iterations. Second, the residual Jacobian computation, which must be performed for all keypoints in both old and new segments, remains unaffected by the surface parameter reutilization method.

\begin{table}[]
	\small\sf\centering
	\caption{Ablation study of surface parameter reutilization on accuracy (unit: m).\label{table10}}
	\vspace{-0.2cm}
	\begin{center}
		\begin{threeparttable}
			\begin{tabular}{p{1.6cm}<{\centering}|p{1.4cm}<{\centering}p{1.4cm}<{\centering}|p{1.8cm}<{\centering}}
				\toprule
				& Base  & Base+ & Metric Error \\ \hline
				$nclt\_1$  & 1.34  & 1.39  & +\,0.05        \\
				$nclt\_2$  & 1.86  & 1.88  & +\,0.02        \\
				$nclt\_3$  & 2.09  & 2.22  & +\,0.13        \\
				$nclt\_4$  & 1.86  & 1.72  & -\,0.14        \\
				$nclt\_5$  & 1.74  & 1.80  & +\,0.06        \\
				$nclt\_6$  & 2.37  & 2.34  & -\,0.03        \\
				$nclt\_7$  & 2.00  & 2.07  & +\,0.07        \\
				$nclt\_8$  & 2.73  & 2.64  & -\,0.09        \\
				$nclt\_9$  & 2.28  & 2.39  & +\,0.11        \\
				$nclt\_10$ & 1.62  & 1.68  & +\,0.06        \\
				$nclt\_11$ & 1.63  & 1.68  & +\,0.05        \\
				$utbm\_1$  & 6.73  & 7.03  & +\,0.30        \\
				$utbm\_2$  & 15.60 & 15.72 & +\,0.12        \\
				$utbm\_3$  & 9.16  & 9.99  & +\,0.83        \\
				$utbm\_4$  & 12.75 & 12.23 & -\,0.52        \\
				$utbm\_5$  & 8.71  & 8.64  & -\,0.07        \\
				$ulhk\_1$  & 0.99  & 1.04  & +\,0.05        \\
				$ulhk\_2$  & 3.14  & 3.22  & +\,0.08        \\
				$kaist\_1$ & 1.15  & 1.11  & -\,0.04        \\
				$kaist\_2$ & 0.91  & 0.84  & -\,0.07        \\
				$kaist\_3$ & 1.42  & 1.37  & -\,0.05        \\ \bottomrule
			\end{tabular}
		\end{threeparttable}
		\begin{tablenotes}
			\footnotesize
			\item[] \textbf{Denotations}: “Base” represents the original SR-LIO with strategic optimizations. “Base+” denotes the subsequent integration of our proposed surface parameter reutilization strategy into the optimized “Base” system.
		\end{tablenotes}
	\end{center}
\end{table}

To rigorously assess the influence of surface parameter reutilization on state estimation accuracy, we conducted comprehensive comparative experiments. As evidenced by the quantitative results presented in Table \ref{table10}, the adoption of surface parameter reutilization yields statistically insignificant variations in estimation accuracy. This empirical finding suggests that the precision and robustness of our state estimation framework exhibit strong resilience to randomness in keypoint distribution patterns. The observed stability further validates our assumption in the Surface Parameter Reutilization section regarding the system's insensitivity to randomness in keypoint distribution patterns.

\subsection{Ablation of Quantized Map Point Management}
\label{Ablation Study of Quantized Map Point Management}

In this section, we conduct an ablation study to systematically evaluate the effects of our proposed quantized map point management method on three critical aspects: memory consumption, computational efficiency, and accuracy. As detailed in the Quantized Map Point Management section, our method employs an index table mapping scheme to compress originally 64-bit double-precision global map points into 8-bit char-type encoded offsets. The final storage overhead of the global map amounts to 7/40 of the original size, accounting for the additional 3$\times$64-bit volume centroid coordinates.

\begin{table}[]
	\small\sf\centering
	\caption{Ablation study of quantized map point management on memory overhead (unit: GB).\label{table11}}
	\vspace{-0.2cm}
	\begin{center}
		\begin{tabular}{p{1.4cm}<{\centering}|p{2.6cm}<{\centering}|p{2.6cm}<{\centering}}
			\toprule
			& \begin{tabular}[c]{@{}c@{}}Standard Map Point\\ Management\end{tabular} & \begin{tabular}[c]{@{}c@{}}Quantized Map Point\\ Management\end{tabular} \\ \hline
			$nclt\_1$  & 0.70                                                                    & 0.36                                                                     \\
			$nclt\_2$  & 0.72                                                                    & 0.37                                                                     \\
			$nclt\_3$  & 0.67                                                                    & 0.37                                                                     \\
			$nclt\_4$  & 0.74                                                                    & 0.39                                                                     \\
			$nclt\_5$  & 0.83                                                                    & 0.41                                                                     \\
			$nclt\_6$  & 0.73                                                                    & 0.38                                                                     \\
			$nclt\_7$  & 0.62                                                                    & 0.36                                                                     \\
			$nclt\_8$  & 0.73                                                                    & 0.39                                                                     \\
			$nclt\_9$  & 0.78                                                                    & 0.37                                                                     \\
			$nclt\_10$ & 0.74                                                                    & 0.39                                                                     \\
			$nclt\_11$ & 0.65                                                                    & 0.38                                                                     \\
			$utbm\_1$  & 0.40                                                                    & 0.25                                                                     \\
			$utbm\_2$  & 0.40                                                                    & 0.25                                                                     \\
			$utbm\_3$  & 0.41                                                                    & 0.23                                                                     \\
			$utbm\_4$  & 0.41                                                                    & 0.25                                                                     \\
			$utbm\_5$  & 0.40                                                                    & 0.24                                                                     \\
			$ulhk\_1$  & 0.26                                                                    & 0.17                                                                     \\
			$ulhk\_2$  & 0.28                                                                    & 0.19                                                                     \\
			$kaist\_1$ & 0.40                                                                    & 0.22                                                                     \\
			$kaist\_2$ & 0.32                                                                    & 0.20                                                                     \\
			$kaist\_3$ & 0.35                                                                    & 0.22                                                                     \\ \bottomrule
		\end{tabular}
	\end{center}
\end{table}

Results in Table \ref{table11} confirm that our quantized map point management method achieves significant memory reduction in LIO operation, though the overall savings fall below the theoretical 7/40 ratio. This discrepancy arises because the system memory footprint comprises not only the global map (where our 8-bit compression applies) but also runtime variables required for LIO’s algorithmic pipeline.

\begin{table*}[]
	\small\sf\centering
	\caption{Ablation study of quantized map point management on time consumption (unit: ms).\label{table12}}
	\vspace{-0.2cm}
	\begin{center}
		\begin{threeparttable}
			\begin{tabular}{p{0.9cm}<{\centering}|p{0.4cm}<{\centering}p{0.4cm}<{\centering}p{0.4cm}<{\centering}p{0.4cm}<{\centering}p{0.4cm}<{\centering}p{0.5cm}<{\centering}|p{0.4cm}<{\centering}p{0.4cm}<{\centering}p{0.4cm}<{\centering}p{0.4cm}<{\centering}p{0.4cm}<{\centering}p{0.5cm}<{\centering}|p{0.4cm}<{\centering}p{0.4cm}<{\centering}p{0.4cm}<{\centering}p{0.4cm}<{\centering}p{0.4cm}<{\centering}p{0.5cm}<{\centering}}
				\toprule
				Platform & \multicolumn{6}{c|}{Intel-Core i9-14900HX}                                                                                                                                                                                                                        & \multicolumn{6}{c|}{Intel Xeon Platinum 8352V}                                                                                                                                                                                                                    & \multicolumn{6}{c}{Raspberry Pi 4B}                                                                                                                                                                                                                                \\ \hline
				Module   & \multicolumn{3}{c|}{Residual Building}                                                                                                    & \multicolumn{3}{c|}{Total}                                                                                            & \multicolumn{3}{c|}{Residual Building}                                                                                                    & \multicolumn{3}{c|}{Total}                                                                                            & \multicolumn{3}{c|}{Residual Building}                                                                                                     & \multicolumn{3}{c}{Total}                                                                                             \\ \hline
				Method   & \begin{tabular}[c]{@{}c@{}}Base\\ +\end{tabular} & \begin{tabular}[c]{@{}c@{}}Base\\ ++\end{tabular} & \multicolumn{1}{c|}{Ours}          & \begin{tabular}[c]{@{}c@{}}Base\\ +\end{tabular} & \begin{tabular}[c]{@{}c@{}}Base\\ ++\end{tabular} & Ours           & \begin{tabular}[c]{@{}c@{}}Base\\ +\end{tabular} & \begin{tabular}[c]{@{}c@{}}Base\\ ++\end{tabular} & \multicolumn{1}{c|}{Ours}          & \begin{tabular}[c]{@{}c@{}}Base\\ +\end{tabular} & \begin{tabular}[c]{@{}c@{}}Base\\ ++\end{tabular} & Ours           & \begin{tabular}[c]{@{}c@{}}Base\\ +\end{tabular} & \begin{tabular}[c]{@{}c@{}}Base\\ ++\end{tabular} & \multicolumn{1}{c|}{Ours}           & \begin{tabular}[c]{@{}c@{}}Base\\ +\end{tabular} & \begin{tabular}[c]{@{}c@{}}Base\\ ++\end{tabular} & Ours           \\ \hline
				$nclt\_1$  & 5.2                                             & 6.2                                              & \multicolumn{1}{c|}{\textbf{4.9}} & 17.0                                            & 17.6                                             & \textbf{15.1} & 6.5                                             & 9.1                                              & \multicolumn{1}{c|}{\textbf{6.1}} & 24.2                                            & 26.2                                             & \textbf{20.5} & 12.6                                            & 15.2                                             & \multicolumn{1}{c|}{\textbf{11.0}} & 50.4                                            & 52.6                                             & \textbf{45.1} \\
				$nclt\_2$  & 5.2                                             & 6.3                                              & \multicolumn{1}{c|}{\textbf{5.0}} & 16.8                                            & 17.5                                             & \textbf{14.9} & 6.6                                             & 9.3                                              & \multicolumn{1}{c|}{\textbf{6.1}} & 23.8                                            & 25.9                                             & \textbf{19.9} & 12.3                                            & 15.0                                             & \multicolumn{1}{c|}{\textbf{10.6}} & 45.9                                            & 49.7                                             & \textbf{40.9} \\
				$nclt\_3$  & 5.2                                             & 6.3                                              & \multicolumn{1}{c|}{\textbf{5.0}} & 17.2                                            & 18.2                                             & \textbf{15.6} & 6.5                                             & 9.2                                              & \multicolumn{1}{c|}{\textbf{6.1}} & 24.3                                            & 26.7                                             & \textbf{20.9} & 12.6                                            & 15.5                                             & \multicolumn{1}{c|}{\textbf{11.0}} & 48.0                                            & 51.0                                             & \textbf{42.3} \\
				$nclt\_4$  & 5.3                                             & 6.2                                              & \multicolumn{1}{c|}{\textbf{4.9}} & 17.6                                            & 18.1                                             & \textbf{15.6} & 6.6                                             & 9.2                                              & \multicolumn{1}{c|}{\textbf{6.1}} & 24.7                                            & 27.0                                             & \textbf{21.0} & 12.9                                            & 15.6                                             & \multicolumn{1}{c|}{\textbf{11.2}} & 51.2                                            & 54.1                                             & \textbf{45.8} \\
				$nclt\_5$  & 5.2                                             & 6.1                                              & \multicolumn{1}{c|}{\textbf{4.5}} & 18.5                                            & 19.1                                             & \textbf{16.4} & 6.8                                             & 9.3                                              & \multicolumn{1}{c|}{\textbf{6.1}} & 26.5                                            & 29.0                                             & \textbf{22.2} & 13.5                                            & 16.3                                             & \multicolumn{1}{c|}{\textbf{11.5}} & 54.8                                            & 58.3                                             & \textbf{48.5} \\
				$nclt\_6$  & 5.2                                             & 6.2                                              & \multicolumn{1}{c|}{\textbf{4.9}} & 17.9                                            & 18.5                                             & \textbf{16.0} & 6.7                                             & 9.4                                              & \multicolumn{1}{c|}{\textbf{6.1}} & 25.4                                            & 28.0                                             & \textbf{21.2} & 12.8                                            & 15.7                                             & \multicolumn{1}{c|}{\textbf{11.1}} & 50.7                                            & 53.8                                             & \textbf{45.1} \\
				$nclt\_7$  & 5.1                                             & 6.1                                              & \multicolumn{1}{c|}{\textbf{4.8}} & 16.9                                            & 17.5                                             & \textbf{15.2} & 6.6                                             & 9.1                                              & \multicolumn{1}{c|}{\textbf{6.0}} & 24.4                                            & 26.1                                             & \textbf{20.2} & 15.4                                            & 15.9                                             & \multicolumn{1}{c|}{\textbf{10.7}} & 47.7                                            & 51.4                                             & \textbf{42.2} \\
				$nclt\_8$  & 4.9                                             & 5.9                                              & \multicolumn{1}{c|}{\textbf{4.6}} & 19.5                                            & 20.6                                             & \textbf{17.7} & 7.2                                             & 9.9                                              & \multicolumn{1}{c|}{\textbf{6.6}} & 29.4                                            & 33.2                                             & \textbf{25.1} & 13.2                                            & 16.6                                             & \multicolumn{1}{c|}{\textbf{11.3}} & 52.1                                            & 57.2                                             & \textbf{47.3} \\
				$nclt\_9$  & 5.1                                             & 6.1                                              & \multicolumn{1}{c|}{\textbf{4.8}} & 17.6                                            & 18.4                                             & \textbf{15.9} & 6.6                                             & 9.2                                              & \multicolumn{1}{c|}{\textbf{6.1}} & 25.2                                            & 27.5                                             & \textbf{21.0} & 12.8                                            & 15.6                                             & \multicolumn{1}{c|}{\textbf{11.1}} & 52.2                                            & 54.5                                             & \textbf{46.2} \\
				$nclt\_10$ & 5.1                                             & 6.2                                              & \multicolumn{1}{c|}{\textbf{4.8}} & 17.2                                            & 17.8                                             & \textbf{15.4} & 6.8                                             & 9.1                                              & \multicolumn{1}{c|}{\textbf{5.9}} & 24.7                                            & 26.1                                             & \textbf{20.1} & 12.9                                            & 16.1                                             & \multicolumn{1}{c|}{\textbf{11.0}} & 49.4                                            & 53.1                                             & \textbf{44.0} \\
				$nclt\_11$ & 5.0                                             & 6.0                                              & \multicolumn{1}{c|}{\textbf{4.7}} & 15.6                                            & 16.3                                             & \textbf{13.9} & 6.1                                             & 8.5                                              & \multicolumn{1}{c|}{\textbf{5.6}} & 21.9                                            & 24.1                                             & \textbf{18.2} & 11.1                                            & 13.6                                             & \multicolumn{1}{c|}{\textbf{9.7}}  & 41.5                                            & 44.0                                             & \textbf{36.7} \\
				$utbm\_1$  & 4.3                                             & 5.0                                              & \multicolumn{1}{c|}{\textbf{4.1}} & 13.6                                            & 14.1                                             & \textbf{12.7} & 5.5                                             & 6.6                                              & \multicolumn{1}{c|}{\textbf{4.8}} & 20.0                                            & 20.5                                             & \textbf{18.0} & 10.2                                            & 11.9                                             & \multicolumn{1}{c|}{\textbf{9.2}}  & 49.0                                            & 50.1                                             & \textbf{43.9} \\
				$utbm\_2$  & 4.7                                             & 5.3                                              & \multicolumn{1}{c|}{\textbf{4.4}} & 15.0                                            & 15.4                                             & \textbf{13.9} & 5.7                                             & 6.8                                              & \multicolumn{1}{c|}{\textbf{5.0}} & 20.6                                            & 21.3                                             & \textbf{18.6} & 10.5                                            & 12.2                                             & \multicolumn{1}{c|}{\textbf{9.5}}  & 49.4                                            & 49.8                                             & \textbf{43.5} \\
				$utbm\_3$  & 4.6                                             & 5.2                                              & \multicolumn{1}{c|}{\textbf{4.3}} & 14.2                                            & 14.8                                             & \textbf{13.4} & 5.6                                             & 6.7                                              & \multicolumn{1}{c|}{\textbf{5.0}} & 19.9                                            & 20.5                                             & \textbf{17.9} & 10.3                                            & 12.3                                             & \multicolumn{1}{c|}{\textbf{9.4}}  & 48.9                                            & 49.4                                             & \textbf{42.3} \\
				$utbm\_4$  & 4.4                                             & 5.2                                              & \multicolumn{1}{c|}{\textbf{4.2}} & 13.8                                            & 14.3                                             & \textbf{12.8} & 5.5                                             & 6.7                                              & \multicolumn{1}{c|}{\textbf{4.9}} & 19.9                                            & 20.7                                             & \textbf{18.0} & 10.2                                            & 12.0                                             & \multicolumn{1}{c|}{\textbf{9.2}}  & 48.9                                            & 49.2                                             & \textbf{43.4} \\
				$utbm\_5$  & 4.1                                             & 4.7                                              & \multicolumn{1}{c|}{\textbf{3.8}} & 13.1                                            & 13.5                                             & \textbf{12.0} & 5.5                                             & 6.5                                              & \multicolumn{1}{c|}{\textbf{4.8}} & 19.6                                            & 20.0                                             & \textbf{17.7} & 10.0                                            & 11.4                                             & \multicolumn{1}{c|}{\textbf{9.0}}  & 45.6                                            & 46.4                                             & \textbf{41.4} \\
				$ulhk\_1$  & 3.4                                             & 4.2                                              & \multicolumn{1}{c|}{\textbf{3.3}} & 13.0                                            & 13.3                                             & \textbf{12.5} & 4.8                                             & 6.7                                              & \multicolumn{1}{c|}{\textbf{4.7}} & 21.3                                            & 23.1                                             & \textbf{20.1} & 9.8                                             & 11.8                                             & \multicolumn{1}{c|}{\textbf{8.7}}  & 50.1                                            & 50.9                                             & \textbf{45.5} \\
				$ulhk\_2$  & 3.4                                             & 4.1                                              & \multicolumn{1}{c|}{\textbf{3.2}} & 12.8                                            & 13.2                                             & \textbf{12.4} & 4.6                                             & 6.5                                              & \multicolumn{1}{c|}{\textbf{4.3}} & 21.1                                            & 23.2                                             & \textbf{20.3} & 10.0                                            & 11.9                                             & \multicolumn{1}{c|}{\textbf{9.3}}  & 51.6                                            & 51.9                                             & \textbf{46.1} \\
				$kaist\_1$ & 5.2                                             & 6.1                                              & \multicolumn{1}{c|}{\textbf{4.9}} & 12.9                                            & 13.3                                             & \textbf{11.8} & 6.0                                             & 7.5                                              & \multicolumn{1}{c|}{\textbf{5.5}} & 16.2                                            & 17.6                                             & \textbf{14.9} & 11.1                                            & 13.3                                             & \multicolumn{1}{c|}{\textbf{10.4}} & 36.5                                            & 37.7                                             & \textbf{32.8} \\
				$kaist\_2$ & 5.1                                             & 5.9                                              & \multicolumn{1}{c|}{\textbf{4.9}} & 13.6                                            & 14.3                                             & \textbf{11.7} & 5.9                                             & 7.4                                              & \multicolumn{1}{c|}{\textbf{5.4}} & 18.0                                            & 18.9                                             & \textbf{15.8} & 10.6                                            & 12.7                                             & \multicolumn{1}{c|}{\textbf{9.9}}  & 37.9                                            & 38.9                                             & \textbf{33.1} \\
				$kaist\_3$ & 5.2                                             & 6.2                                              & \multicolumn{1}{c|}{\textbf{4.8}} & 11.1                                            & 11.1                                             & \textbf{9.6}  & 6.5                                             & 7.5                                              & \multicolumn{1}{c|}{\textbf{5.5}} & 14.3                                            & 14.5                                             & \textbf{12.1} & 11.1                                            & 11.3                                             & \multicolumn{1}{c|}{\textbf{8.6}}  & 25.8                                            & 29.1                                             & \textbf{23.5} \\ \bottomrule
			\end{tabular}
		\end{threeparttable}
		\begin{tablenotes}
			\footnotesize
			\item[] \textbf{Denotations}: “Base+” denotes the enhanced SR-LIO system incorporating both  additional strategic optimization and our proposed surface parameter reutilization method. “Base++” extends the “Base+” system by incorporating quantized map point management, where nearest neighbor search operations are performed in double-precision domain after decoding the compressed map points, including full-precision Euclidean distance calculation and sorting. “Residual Building” refers to the computational time for constructing point-to-plane residuals in each state update iteration, comprising nearest neighbor search, surface fitting, and residual Jacobian computation. It is executed $m$ times per reconstructed sweep (where $1 \le m \le 5$). “Total” indicates the complete processing time per reconstructed sweep.
		\end{tablenotes}
	\end{center}
\end{table*}

Results in Table \ref{table12} reveal two key findings regarding computational efficiency: 1) When maintaining SR-LIO's original nearest neighbor search processing but utilizing quantized map point management (denoted as “Base++”), the computational overhead increases significantly due to the mandatory decoding operations before each Euclidean distance calculation. 2) By implementing our proposed integer-domain nearest neighbor search, we not only recover the efficiency loss but achieve further computational improvements over “Base+”, ultimately enabling consistent 20\,Hz state estimation across all sequences. The same as the Ablation of Surface Parameter Reutilization section, we evaluate the time overhead of “Residual Building” module to mitigate the potential inflation of timing measurements caused by the high-frequency execution of timing functions. The “Residual Building Module” encapsulates all computational components where quantized map point management can potentially contribute to efficiency improvements, making it an appropriate metric for assessing the impact of our proposed method.

\begin{table}[]
	\vspace{-0.5cm}
	\small\sf\centering
	\caption{Ablation study of quantized map point management on time consumption of map update module (unit: ms).\label{table13}}
	\vspace{-0.2cm}
	\begin{center}
		\begin{threeparttable}
			\begin{tabular}{c|p{0.5cm}<{\centering}p{0.7cm}<{\centering}|p{0.7cm}<{\centering}p{0.7cm}<{\centering}|p{0.7cm}<{\centering}p{0.7cm}<{\centering}}
				\toprule
				Platform & \multicolumn{2}{c|}{\begin{tabular}[c]{@{}c@{}}Intel-Core\\ i9-14900HX\end{tabular}} & \multicolumn{2}{c|}{\begin{tabular}[c]{@{}c@{}}Intel Xeon\\ Platinum 8352V\end{tabular}} & \multicolumn{2}{c}{\begin{tabular}[c]{@{}c@{}}Raspberry\\ Pi 4B\end{tabular}} \\ \hline
				Method   & Base+                                     & Ours                                     & Base+                                       & Ours                                       & Base+                                  & Ours                                 \\ \hline
				$nclt\_1$  & 0.13                                      & 0.19                                     & 0.20                                        & 0.32                                       & 0.59                                   & 0.72                                 \\
				$nclt\_2$  & 0.10                                      & 0.17                                     & 0.18                                        & 0.28                                       & 0.48                                   & 0.62                                 \\
				$nclt\_3$  & 0.12                                      & 0.18                                     & 0.19                                        & 0.31                                       & 0.56                                   & 0.70                                 \\
				$nclt\_4$  & 0.12                                      & 0.18                                     & 0.19                                        & 0.32                                       & 0.62                                   & 0.79                                 \\
				$nclt\_5$  & 0.10                                      & 0.16                                     & 0.19                                        & 0.29                                       & 0.66                                   & 0.80                                 \\
				$nclt\_6$  & 0.10                                      & 0.17                                     & 0.18                                        & 0.30                                       & 0.68                                   & 0.82                                 \\
				$nclt\_7$  & 0.13                                      & 0.19                                     & 0.23                                        & 0.35                                       & 0.66                                   & 0.82                                 \\
				$nclt\_8$  & 0.10                                      & 0.15                                     & 0.19                                        & 0.30                                       & 0.63                                   & 0.81                                 \\
				$nclt\_9$  & 0.12                                      & 0.18                                     & 0.23                                        & 0.32                                       & 0.72                                   & 0.87                                 \\
				$nclt\_10$ & 0.11                                      & 0.18                                     & 0.19                                        & 0.30                                       & 0.53                                   & 0.66                                 \\
				$nclt\_11$ & 0.11                                      & 0.18                                     & 0.17                                        & 0.29                                       & 0.48                                   & 0.61                                 \\
				$utbm\_1$  & 0.18                                      & 0.28                                     & 0.24                                        & 0.41                                       & 0.71                                   & 0.87                                 \\
				$utbm\_2$  & 0.18                                      & 0.29                                     & 0.25                                        & 0.42                                       & 0.69                                   & 0.84                                 \\
				$utbm\_3$  & 0.21                                      & 0.31                                     & 0.29                                        & 0.45                                       & 0.74                                   & 0.91                                 \\
				$utbm\_4$  & 0.18                                      & 0.29                                     & 0.25                                        & 0.42                                       & 0.75                                   & 0.92                                 \\
				$utbm\_5$  & 0.17                                      & 0.25                                     & 0.24                                        & 0.40                                       & 0.72                                   & 0.86                                 \\
				$ulhk\_1$  & 0.09                                      & 0.17                                     & 0.14                                        & 0.27                                       & 0.43                                   & 0.58                                 \\
				$ulhk\_2$  & 0.18                                      & 0.29                                     & 0.26                                        & 0.52                                       & 0.96                                   & 1.20                                 \\
				$kaist\_1$ & 0.21                                      & 0.33                                     & 0.23                                        & 0.39                                       & 0.71                                   & 0.88                                 \\
				$kaist\_2$ & 0.16                                      & 0.28                                     & 0.19                                        & 0.35                                       & 0.56                                   & 0.75                                 \\
				$kaist\_3$ & 0.11                                      & 0.17                                     & 0.12                                        & 0.20                                       & 0.37                                   & 0.46                                 \\ \bottomrule
			\end{tabular}
		\end{threeparttable}
		\begin{tablenotes}
			\footnotesize
			\item[] \textbf{Denotations}: “Base+” denotes the enhanced SR-LIO system incorporating both  additional strategic optimization and our proposed surface parameter reutilization method.
		\end{tablenotes}
	\end{center}
\end{table}

\begin{table}[]
	\vspace{-0.5cm}
	\small\sf\centering
	\caption{Ablation study of quantized map point management on accuracy (unit: m).\label{table14}}
	\vspace{-0.2cm}
	\begin{center}
		\begin{threeparttable}
			\begin{tabular}{p{1.6cm}<{\centering}|p{1.6cm}<{\centering}p{1.6cm}<{\centering}|p{1.8cm}<{\centering}}
				\toprule
				& Base+ & Ours  & Metric Error \\ \hline
				$nclt\_1$  & 1.39  & 1.34  & -\,0.05        \\
				$nclt\_2$  & 1.88  & 1.56  & -\,0.32        \\
				$nclt\_3$  & 2.22  & 2.25  & +\,0.03        \\
				$nclt\_4$  & 1.72  & 1.54  & -\,0.18        \\
				$nclt\_5$  & 1.80  & 1.92  & +\,0.12        \\
				$nclt\_6$  & 2.34  & 2.24  & -\,0.10        \\
				$nclt\_7$  & 2.07  & 1.96  & -\,0.11        \\
				$nclt\_8$  & 2.64  & 2.44  & -\,0.20        \\
				$nclt\_9$  & 2.39  & 2.35  & -\,0.04        \\
				$nclt\_10$ & 1.68  & 1.63  & -\,0.05        \\
				$nclt\_11$ & 1.68  & 1.54  & -\,0.14        \\
				$utbm\_1$  & 7.03  & 7.29  & +\,0.26        \\
				$utbm\_2$  & 15.72 & 15.25 & -\,0.47        \\
				$utbm\_3$  & 9.99  & 11.31 & +\,1.32        \\
				$utbm\_4$  & 12.23 & 10.46 & -\,1.77        \\
				$utbm\_5$  & 8.64  & 8.48  & -\,0.16        \\
				$ulhk\_1$  & 1.04  & 1.13  & +\,0.09        \\
				$ulhk\_2$  & 3.22  & 3.18  & -\,0.04        \\
				$kaist\_1$ & 1.11  & 1.12  & +\,0.01        \\
				$kaist\_2$ & 0.84  & 0.87  & +\,0.03        \\
				$kaist\_3$ & 1.37  & 1.36  & -\,0.01        \\ \bottomrule
			\end{tabular}
		\end{threeparttable}
		\begin{tablenotes}
			\footnotesize
			\item[] \textbf{Denotations}: “Base+” denotes the enhanced SR-LIO system incorporating both  additional strategic optimization and our proposed surface parameter reutilization method.
		\end{tablenotes}
	\end{center}
\end{table}

As discussed in the Map Update section, the quantized map management scheme introduces additional encoding/decoding operations during the map update phase. We conduct a systematic evaluation to quantify the associated computational overhead. The results in Table \ref{table13} demonstrate that while these operations indeed increase the computational load of map update, the absolute overhead remains negligible due to the inherently low baseline computational cost of this module.

Experimental results in Table \ref{table14} demonstrate that map quantization induces negligible effects on state estimation accuracy across all sequences except $utbm\_3$. This observation validates our assumption in the Quantized Map Point Management section, confirming that the 3.9\,mm-quantization resolution introduces clinically insignificant precision loss for LIO systems.

\subsection{Time Consumption of System Modules}
\label{Time Consumption of System Modules}

\begin{table*}[]
	\small\sf\centering
	\caption{Time consumption of system modules (unit: ms).\label{table15}}
	\vspace{-0.2cm}
	\begin{center}
	\begin{threeparttable}
		\begin{tabular}{c|p{0.9cm}<{\centering}p{0.9cm}<{\centering}cp{0.9cm}<{\centering}|p{0.9cm}<{\centering}p{0.9cm}<{\centering}cp{0.9cm}<{\centering}|p{0.9cm}<{\centering}p{0.9cm}<{\centering}cp{0.9cm}<{\centering}}
			\toprule
			Platform & \multicolumn{4}{c|}{Intel-Core i9-14900HX}                                                                                                                                             & \multicolumn{4}{c|}{Intel Xeon Platinum 8352V}                                                                                                                                         & \multicolumn{4}{c}{Raspberry Pi 4B}                                                                                                                                                    \\ \hline
			Module   & (1) & (2) & (3) & Total & (1) & (2) & (3) & Total & (1) & (2) & (3) & Total \\ \hline
			$nclt\_1$  & 0.9                                                       & 11.0                                                      & 0.2                                                 & 15.1 & 1.4                                                       & 13.9                                                      & 0.3                                                 & 20.5 & 5.8                                                       & 26.2                                                      & 0.7                                                 & 45.1 \\
			$nclt\_2$  & 0.9                                                       & 10.9                                                      & 0.2                                                 & 14.9 & 1.4                                                       & 13.5                                                      & 0.3                                                 & 19.9 & 5.3                                                       & 24.3                                                      & 0.6                                                 & 40.9 \\
			$nclt\_3$  & 0.9                                                       & 11.4                                                      & 0.2                                                 & 15.6 & 1.4                                                       & 14.3                                                      & 0.3                                                 & 20.9 & 4.7                                                       & 26.7                                                      & 0.7                                                 & 42.3 \\
			$nclt\_4$  & 0.8                                                       & 11.4                                                      & 0.2                                                 & 15.6 & 1.4                                                       & 14.5                                                      & 0.3                                                 & 21.0 & 5.3                                                       & 27.9                                                      & 0.8                                                 & 45.8 \\
			$nclt\_5$  & 0.9                                                       & 12.0                                                      & 0.2                                                 & 16.4 & 1.4                                                       & 15.7                                                      & 0.3                                                 & 22.2 & 5.0                                                       & 31.0                                                      & 0.8                                                 & 48.5 \\
			$nclt\_6$  & 0.9                                                       & 11.6                                                      & 0.2                                                 & 16.0 & 1.4                                                       & 14.6                                                      & 0.3                                                 & 21.2 & 5.0                                                       & 27.9                                                      & 0.8                                                 & 45.1 \\
			$nclt\_7$  & 1.0                                                       & 10.9                                                      & 0.2                                                 & 15.2 & 1.4                                                       & 13.7                                                      & 0.4                                                 & 20.2 & 5.2                                                       & 25.0                                                      & 0.8                                                 & 42.2 \\
			$nclt\_8$  & 0.9                                                       & 13.3                                                      & 0.2                                                 & 17.7 & 1.4                                                       & 18.7                                                      & 0.3                                                 & 25.1 & 5.3                                                       & 28.8                                                      & 0.8                                                 & 47.3 \\
			$nclt\_9$  & 0.9                                                       & 11.4                                                      & 0.2                                                 & 15.9 & 1.5                                                       & 14.5                                                      & 0.3                                                 & 21.0 & 5.4                                                       & 27.9                                                      & 0.9                                                 & 46.2 \\
			$nclt\_10$ & 1.0                                                       & 11.0                                                      & 0.2                                                 & 15.4 & 1.4                                                       & 13.7                                                      & 0.3                                                 & 20.1 & 5.6                                                       & 26.0                                                      & 0.7                                                 & 44.0 \\
			$nclt\_11$ & 0.9                                                       & 9.9                                                       & 0.2                                                 & 13.9 & 1.3                                                       & 12.2                                                      & 0.3                                                 & 18.2 & 4.9                                                       & 21.4                                                      & 0.6                                                 & 36.7 \\
			$utbm\_1$  & 1.8                                                       & 7.6                                                       & 0.3                                                 & 12.7 & 3.3                                                       & 9.3                                                       & 0.4                                                 & 18.0 & 13.4                                                      & 18.7                                                      & 0.9                                                 & 43.9 \\
			$utbm\_2$  & 1.7                                                       & 8.5                                                       & 0.3                                                 & 13.9 & 3.1                                                       & 10.1                                                      & 0.4                                                 & 18.6 & 12.7                                                      & 19.7                                                      & 0.8                                                 & 43.5 \\
			$utbm\_3$  & 1.7                                                       & 8.2                                                       & 0.3                                                 & 13.4 & 3.0                                                       & 9.8                                                       & 0.5                                                 & 17.9 & 12.4                                                      & 20.0                                                      & 0.9                                                 & 42.3 \\
			$utbm\_4$  & 1.8                                                       & 7.6                                                       & 0.3                                                 & 12.8 & 3.3                                                       & 9.2                                                       & 0.4                                                 & 18.0 & 13.2                                                      & 18.4                                                      & 0.9                                                 & 43.4 \\
			$utbm\_5$  & 2.0                                                       & 7.0                                                       & 0.3                                                 & 12.0 & 3.6                                                       & 9.1                                                       & 0.4                                                 & 17.7 & 13.3                                                      & 17.7                                                      & 0.9                                                 & 41.4 \\
			$ulhk\_1$  & 4.1                                                       & 5.7                                                       & 0.2                                                 & 12.5 & 6.2                                                       & 8.6                                                       & 0.3                                                 & 20.1 & 17.6                                                      & 16.4                                                      & 0.6                                                 & 45.5 \\
			$ulhk\_2$  & 3.2                                                       & 5.8                                                       & 0.3                                                 & 12.4 & 5.2                                                       & 10.6                                                      & 0.5                                                 & 20.3 & 20.0                                                      & 19.6                                                      & 1.2                                                 & 46.1 \\
			$kaist\_1$ & 0.5                                                       & 8.4                                                       & 0.3                                                 & 11.8 & 0.7                                                       & 9.7                                                       & 0.4                                                 & 14.9 & 4.1                                                       & 19.3                                                      & 0.9                                                 & 32.8 \\
			$kaist\_2$ & 0.5                                                       & 9.4                                                       & 0.3                                                 & 11.7 & 0.7                                                       & 10.8                                                      & 0.4                                                 & 15.8 & 4.2                                                       & 20.5                                                      & 0.8                                                 & 33.1 \\
			$kaist\_3$ & 0.5                                                       & 6.9                                                       & 0.2                                                 & 9.6  & 0.6                                                       & 8.1                                                       & 0.2                                                 & 12.1 & 3.3                                                       & 13.7                                                      & 0.5                                                 & 23.5 \\ \bottomrule
		\end{tabular}
	\end{threeparttable}
	\begin{tablenotes}
	\footnotesize
	\item[] \textbf{Denotations}: (1) point cloud processing; (2) state estimation; (3) map update.
	\end{tablenotes}
	\end{center}
\end{table*}

We conduct a comprehensive runtime evaluation across all sequences, measuring the computational overhead of three core modules: 1) Point cloud processing: Raw point cloud data down-sampling and sweep reconstruction; 2) State estimation: State prediction and iterative state update (excluding map update); 3) Map update: New point registration and far-point removal. Table \ref{table15} presents the processing time distribution across above modules for individual reconstructed sweeps. Our system demonstrates stable 20\,Hz real-time performance across three distinct computational platforms with varying processing capabilities.

\subsection{Evaluation on Custom-Built Sensor System}
\label{Evaluation on Custom-Built Sensor System}

\begin{figure}
	\setlength{\fboxsep}{0pt}%
	\setlength{\fboxrule}{0pt}%
	\begin{center}
		\includegraphics[width=0.97\linewidth]{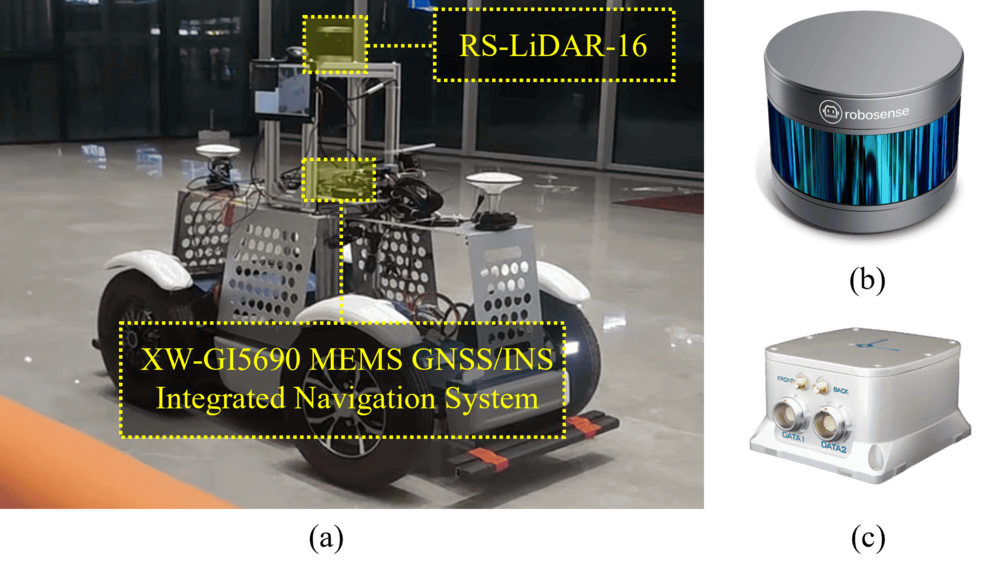}
	\end{center}
	\caption{(a) Sensor configuration for experiments on our own platform. (b) A RoboSense RS-LiDAR-16 LiDAR sensor is rigidly synchronized with (c) a XW-GI5690 MEMS GNSS/INS integrated navigation system through hardware triggering for multimodal data collection.}
	\label{fig10}
\end{figure}

To validate the effectiveness of SR-LIO++ in real-world scenarios, we deployed our proprietary hardware platform (Fig. \ref{fig10}) for collecting experimental data through field tests, followed by offline evaluations across three hardware platforms with distinct computational capabilities. The platform equips a Robosense RS-LiDAR-16 3D LiDAR and a StarNet XW-GI5690 MEMS GNSS/INS integrated navigation system. The RS-LiDAR-16 acquires point cloud data at 10\,Hz, while the integrated navigation system operates at 100\,Hz, with both devices synchronized through hardware triggering for precise temporal alignment.

We evaluated SR-LIO++ using data collected from urban environments. The global positioning information from the integrated navigation system served as the reference for assessing SR-LIO++'s localization accuracy. A total of five sequences were recorded. However, due to inaccuracies in the integrated navigation output under heavy foliage occlusion and inside buildings, comparative analysis was conducted only on four sequences (i.e., Seq. 1$\sim$4) where such interference was absent.

\begin{figure}
	\setlength{\fboxsep}{0pt}%
	\setlength{\fboxrule}{0pt}%
	\begin{center}
		\includegraphics[width=0.97\linewidth]{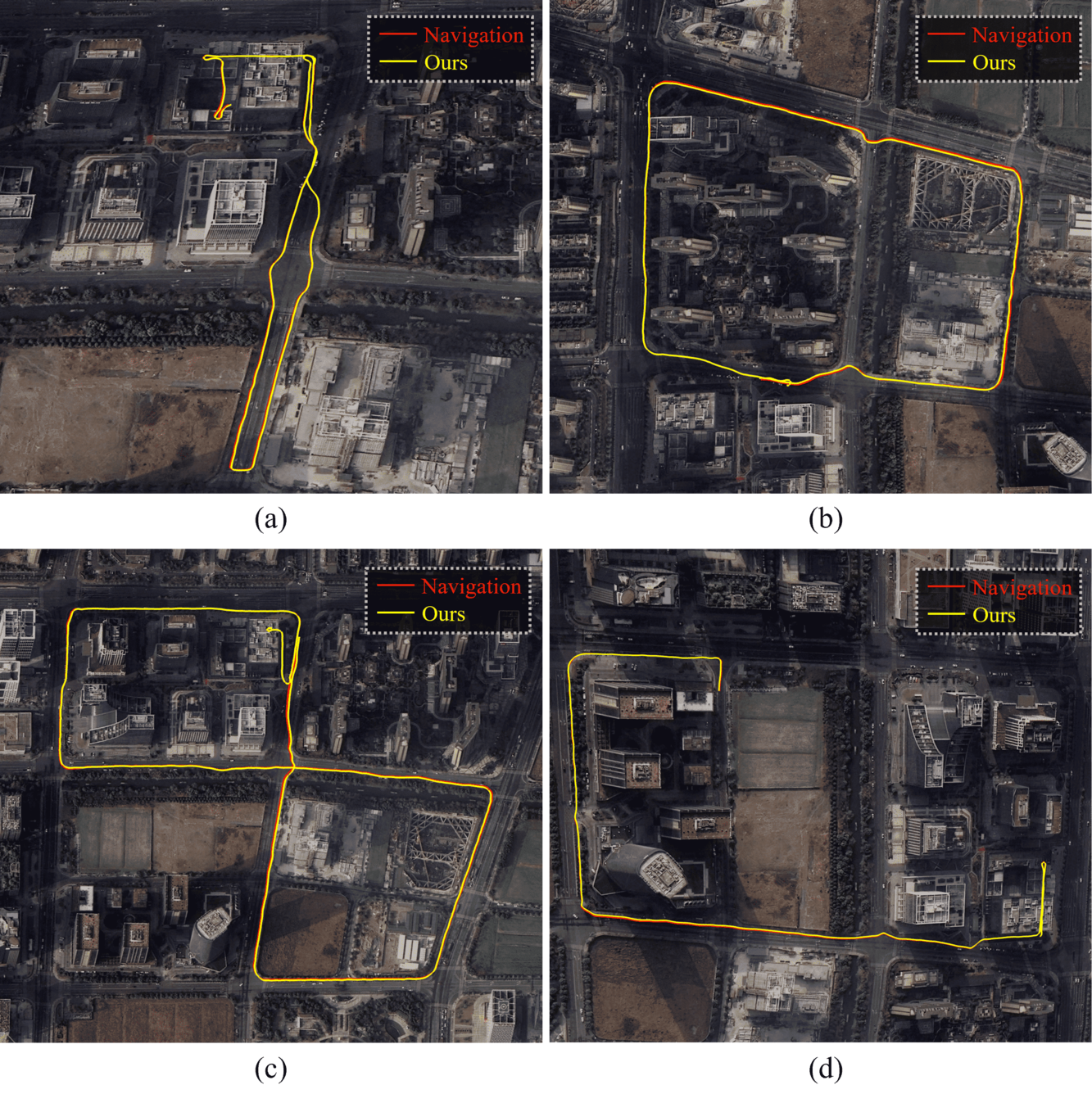}
	\end{center}
	\caption{The comparative visualization between SR-LIO++'s estimated trajectories and the navigation solutions on Google Earth for Seq. 1$\sim$4.}
	\label{fig11}
\end{figure}

\begin{figure}
	\setlength{\fboxsep}{0pt}%
	\setlength{\fboxrule}{0pt}%
	\begin{center}
		\includegraphics[width=0.97\linewidth]{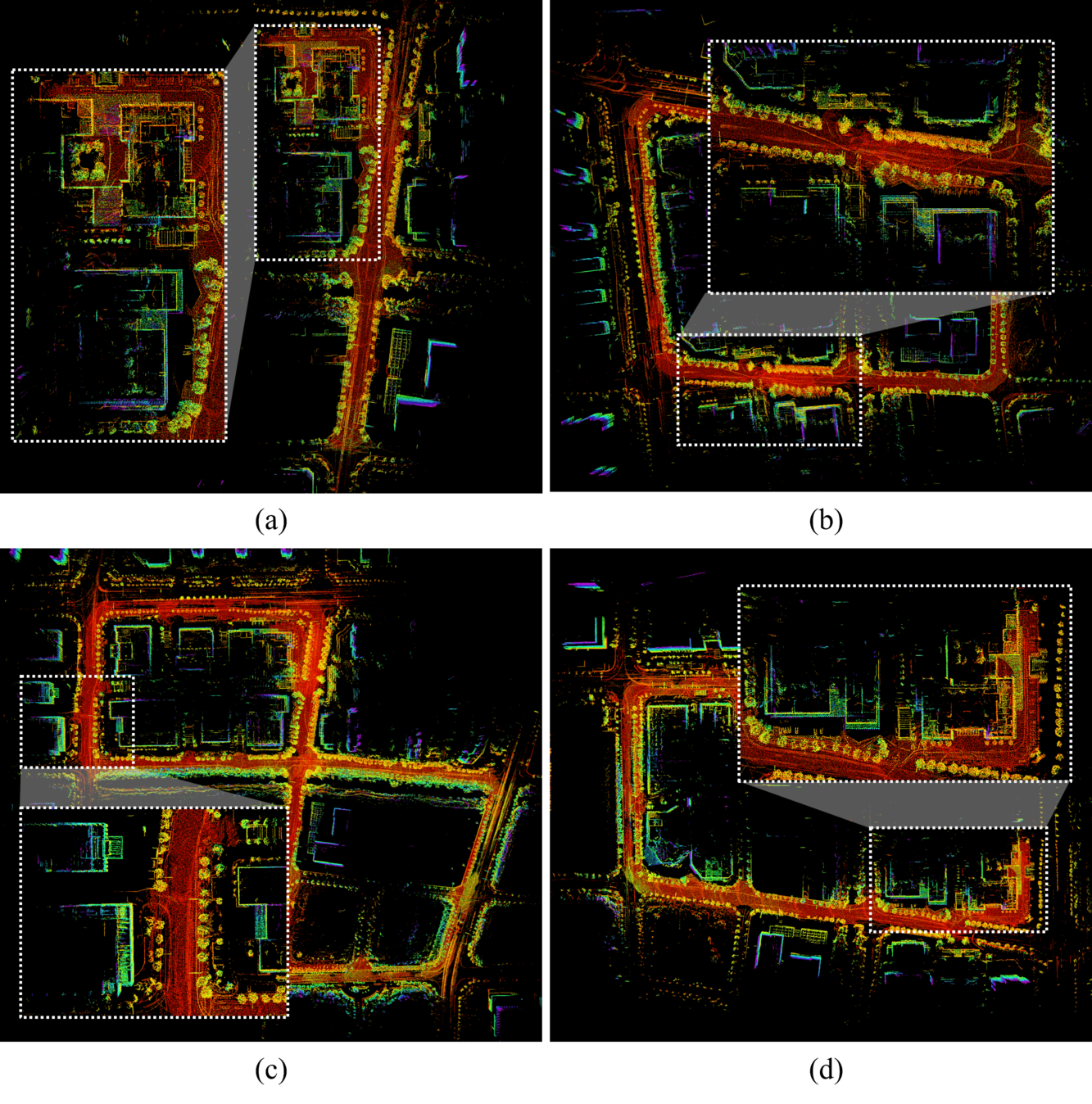}
	\end{center}
	\caption{The visualization of reconstructed point cloud map for Seq. 1$\sim$4.}
	\label{fig12}
\end{figure}

\begin{table}[]
	\small\sf\centering
	\caption{Time consumption on self-collected dataset (unit: ms).\label{table16}}
	\vspace{-0.2cm}
	\begin{center}
		\begin{tabular}{p{1.3cm}<{\centering}|p{1.6cm}<{\centering}|p{2.0cm}<{\centering}|p{1.6cm}<{\centering}}
			\toprule
			Platform & \begin{tabular}[c]{@{}c@{}}Intel-Core\\ i9-14900HX\end{tabular} & \begin{tabular}[c]{@{}c@{}}Intel Xeon\\ Platinum 8352V\end{tabular} & \begin{tabular}[c]{@{}c@{}}Raspberry\\ Pi 4B\end{tabular} \\ \hline
			Seq. 1   & 14.6                                                           & 18.0                                                               & 42.3                                                     \\
			Seq. 2   & 13.4                                                           & 16.5                                                               & 40.6                                                     \\
			Seq. 3   & 14.6                                                           & 16.7                                                               & 41.3                                                     \\
			Seq. 4   & 16.8                                                           & 23.3                                                               & 47.4                                                     \\
			Seq. 5   & 13.8                                                           & 17.2                                                               & 45.8                                                     \\ \bottomrule
		\end{tabular}
	\end{center}
\end{table}

Fig. \ref{fig11} (a)–(d) present the visualization of SR-LIO++'s estimated trajectories overlaid with the navigation solutions on Google Earth. The near-perfect alignment between the trajectories demonstrates that SR-LIO++ achieves high localization accuracy even in real-world environments. Furthermore, Fig. \ref{fig12} (a)–(d) display the globally consistent point cloud maps corresponding to the four sequences in Fig. \ref{fig11}, showing high-quality reconstruction without noticeable ghosting artifacts. For Seq. 5 containing indoor segments, we provide the visualization results in Fig. \ref{fig1}, while omitting the comparison with the integrated navigation system's global positioning data due to its unreliability in GNSS-denied environments. The “supplementary video” further demonstrates the performance on Seq. 1$\sim$Seq. 5. Table \ref{table16} documents the average computational across five self-collected sequences on three computing platforms.

\section{Conclusion}
\label{Conclusion}

In this work, we have introduced SR-LIO++, an advanced LiDAR-Inertial Odometry system that addresses the critical challenge of low-frequency 3D LiDAR data acquisition while maintaining robust real-time performance on computationally constrained platforms. Building upon the sweep reconstruction method, we incorporate an intelligent caching mechanism for intermediate surface parameters. This architecture effectively eliminates redundant processing of overlapping segments in consecutive reconstructed sweeps, thereby breaking the traditional linear scaling between computational load and output frequency enhancement. Then, we present a quantization scheme for map point management utilizing index table mapping. Our approach achieves a remarkable reduction in memory requirements by transitioning from 64-bit floating-point to compact 8-bit char formats for 3D point storage, while maintaining full reconstruction capability. We also have reformulated the computationally expensive nearest neighbor search operations by transforming the underlying Euclidean distance calculations from 64-bit floating-point arithmetic to optimized 16-bit/32-bit integer domains. This innovation yields substantial improvements in both computational efficiency and memory utilization. Comprehensive experimental evaluation across diverse hardware platforms (including embedded systems) and multiple public datasets demonstrates that SR-LIO++ consistently maintains state-of-the-art localization accuracy while achieving unprecedented computational efficiency. Most notably, the system successfully attains 20\,Hz real-time state estimation on a Raspberry Pi 4B single-board computer, representing a significant advancement in frequency-enhanced LIO systems for resource-constrained computing platforms. Future work will focus on deploying SR-LIO++ in LiDAR-based autonomous exploration framework.

\begin{acks}
This research was partially conducted by ACCESS – AI Chip Center for Emerging Smart Systems, supported by the InnoHK initiative of the Innovation and Technology Commission of the Hong Kong Special Administrative Region Government, and was partially conducted by National Natural Science Foundation of China (62122029, 62472184) and the Fundamental Research Funds for the Central Universities.
\end{acks}


\begin{thebibliography}{99}

\bibitem[Zhang and Singh, 2014]{zhang2014loam}
Zhang~J and Singh~S (2014) LOAM: Lidar odometry and mapping in real-time.
\textit{Proceedings of Robotics: Science and Systems}, Berkeley, CA, Vol.~2, No.~9, pp.~1--9.

\bibitem[Shan and Englot, 2018]{shan2018lego}
Shan~T and Englot~B (2018) Lego-loam: Lightweight and ground-optimized lidar odometry and mapping on variable terrain.
\textit{2018 IEEE/RSJ International Conference on Intelligent Robots and Systems (IROS)}, Madrid, Spain, pp.~4758--4765.

\bibitem[Wang et al., 2020]{wang2020intensity}
Wang~H, Wang~C and Xie~L (2020) Intensity scan context: Coding intensity and geometry relations for loop closure detection.
\textit{2020 IEEE International Conference on Robotics and Automation (ICRA)}, Paris, France, pp.~2095--2101.

\bibitem[Deschaud, 2018]{deschaud2018imls}
Deschaud~JE (2018) IMLS-SLAM: Scan-to-model matching based on 3D data.
\textit{2018 IEEE International Conference on Robotics and Automation (ICRA)}, Brisbane, Australia, pp.~2480--2485.

\bibitem[Dellenbach et al., 2022]{dellenbach2022ct}
Dellenbach~P, Deschaud~JE, Jacquet~B and Goulette~F (2022) Ct-icp: Real-time elastic lidar odometry with loop closure.
\textit{2022 International Conference on Robotics and Automation (ICRA)}, Philadelphia, USA, pp.~5580--5586.

\bibitem[Li et al., 2021]{li2021towards}
Li~K, Li~M and Hanebeck~UD (2021) Towards high-performance solid-state-lidar-inertial odometry and mapping.
\textit{IEEE Robotics and Automation Letters}, Vol.~6, No.~3, pp.~5167--5174.

\bibitem[Ye et al., 2019]{ye2019tightly}
Ye~H, Chen~Y and Liu~M (2019) Tightly coupled 3d lidar inertial odometry and mapping.
\textit{2019 International Conference on Robotics and Automation (ICRA)}, Montreal, Canada, pp.~3144--3150.

\bibitem[Qin et al., 2020]{qin2020lins}
Qin~C, Ye~H, Pranata~CE, Han~J, Zhang~S and Liu~M (2020) Lins: A lidar-inertial state estimator for robust and efficient navigation.
\textit{2020 IEEE International Conference on Robotics and Automation (ICRA)}, Paris, France, pp.~8899--8906.

\bibitem[Shan et al., 2020]{shan2020lio}
Shan~T, Englot~B, Meyers~D, Wang~W, Ratti~C and Rus~D (2020) Lio-sam: Tightly-coupled lidar inertial odometry via smoothing and mapping.
\textit{2020 IEEE/RSJ International Conference on Intelligent Robots and Systems (IROS)}, Las Vegas, USA, pp.~5135--5142.

\bibitem[Xu and Zhang, 2021]{xu2021fast}
Xu~W and Zhang~F (2021) Fast-lio: A fast, robust lidar-inertial odometry package by tightly-coupled iterated kalman filter.
\textit{IEEE Robotics and Automation Letters}, Vol.~6, No.~2, pp.~3317--3324.

\bibitem[Xu et al., 2022]{xu2022fast}
Xu~W, Cai~Y, He~D, Lin~J and Zhang~F (2022) Fast-lio2: Fast direct lidar-inertial odometry.
\textit{IEEE Transactions on Robotics}, Vol.~38, No.~4, pp.~2053--2073.

\bibitem[Chen et al., 2023]{chen2023direct}
Chen~K, Nemiroff~R and Lopez~BT (2023) Direct lidar-inertial odometry: Lightweight lio with continuous-time motion correction.
\textit{2023 IEEE International Conference on Robotics and Automation (ICRA)}, London, UK, pp.~3983--3989.

\bibitem[Behley and Stachniss, 2018]{behley2018efficient}
Behley~J and Stachniss~C (2018) Efficient surfel-based SLAM using 3D laser range data in urban environments.
\textit{Robotics: Science and Systems}, Pittsburgh, USA, Vol.~2018, pp.~59.

\bibitem[Qin et al., 2018]{qin2018vins}
Qin~T, Li~P and Shen~S (2018) Vins-mono: A robust and versatile monocular visual-inertial state estimator.
\textit{IEEE Transactions on Robotics}, Vol.~34, No.~4, pp.~1004--1020.

\bibitem[Geneva et al., 2020]{geneva2020openvins}
Geneva~P, Eckenhoff~K, Lee~W, Yang~Y and Huang~G (2020) Openvins: A research platform for visual-inertial estimation.
\textit{2020 IEEE International Conference on Robotics and Automation (ICRA)}, Paris, France, pp.~4666--4672.

\bibitem[Carlevaris-Bianco et al., 2016]{carlevaris2016university}
Carlevaris-Bianco~N, Ushani~AK and Eustice~RM (2016) University of Michigan North Campus long-term vision and lidar dataset.
\textit{The International Journal of Robotics Research}, Vol.~35, No.~9, pp.~1023--1035.

\bibitem[Yan et al., 2020]{yaneu}
Yan~Z, Sun~L, Krajn{\i}k~T and Ruichek~Y (2020) Eu long-term dataset with multiple sensors for autonomous driving.
\textit{RSJ International Conference on Intelligent Robots and Systems (IROS)}, Las Vegas, USA, pp.~10697--10704.

\bibitem[Wen et al., 2020]{wen2020urbanloco}
Wen~W, Zhou~Y, Zhang~G, Fahandezh-Saadi~S, Bai~X, Zhan~W, Tomizuka~M and Hsu~LT (2020) UrbanLoco: A full sensor suite dataset for mapping and localization in urban scenes.
\textit{2020 IEEE International Conference on Robotics and Automation (ICRA)}, Paris, France, pp.~2310--2316.

\bibitem[Jeong et al., 2019]{jeong2019complex}
Jeong~J, Cho~Y, Shin~YS, Roh~H and Kim~A (2019) Complex urban dataset with multi-level sensors from highly diverse urban environments.
\textit{The International Journal of Robotics Research}, Vol.~38, No.~6, pp.~642--657.

\bibitem[Chen et al., 2024]{chen2024ig}
Chen~Z, Xu~Y, Yuan~S and Xie~L (2024) ig-lio: An incremental gicp-based tightly-coupled lidar-inertial odometry.
\textit{IEEE Robotics and Automation Letters}, Vol.~9, No.~2, pp.~1883--1890.

\bibitem[Bai et al., 2022]{bai2022faster}
Bai~C, Xiao~T, Chen~Y, Wang~H, Zhang~F and Gao~X (2022) Faster-LIO: Lightweight tightly coupled LiDAR-inertial odometry using parallel sparse incremental voxels.
\textit{IEEE Robotics and Automation Letters}, Vol.~7, No.~2, pp.~4861--4868.

\bibitem[Qu et al., 2022]{qu2022llol}
Qu~C, Shivakumar~SS, Liu~W and Taylor~CJ (2022) Llol: Low-latency odometry for spinning lidars.
\textit{2022 International Conference on Robotics and Automation (ICRA)}, Philadelphia, USA, pp.~4149--4155.

\bibitem[He et al., 2023]{he2023point}
He~D, Xu~W, Chen~N, Kong~F, Yuan~C and Zhang~F (2023) Point-LIO: Robust high-bandwidth light detection and ranging inertial odometry.
\textit{Advanced Intelligent Systems}, Vol.~5, No.~7, pp.~2200459.

\bibitem[Yuan et al., 2024a]{yuan2022sr}
Yuan~Z, Lang~F, Xu~T and Yang~X (2024) Sr-lio: Lidar-inertial odometry with sweep reconstruction.
\textit{2024 IEEE/RSJ International Conference on Intelligent Robots and Systems (IROS)}, Abu Dhabi, UAE, pp.~7862--7869.

\bibitem[Yuan et al., 2023a]{yuan2023sdv}
Yuan~Z, Wang~Q, Cheng~K, Hao~T and Yang~X (2023) SDV-LOAM: Semi-direct visual--LiDAR Odometry and mapping.
\textit{IEEE Transactions on Pattern Analysis and Machine Intelligence}, Vol.~45, No.~9, pp.~11203--11220.

\bibitem[Yuan et al., 2024b]{yuan2024sr}
Yuan~Z, Deng~J, Ming~R, Lang~F and Yang~X (2024) Sr-livo: Lidar-inertial-visual odometry and mapping with sweep reconstruction.
\textit{IEEE Robotics and Automation Letters}, Vol.~9, No.~12, pp.~10985--10992.

\bibitem[Yuan et al., 2025]{yuan2025semi}
Yuan~Z, Lang~F, Xu~T, Ming~R, Zhao~C and Yang~X (2025) Semi-elastic LiDAR-inertial odometry.
\textit{2025 IEEE International Conference on Robotics and Automation (ICRA)}, Atlanta, USA, pp.~9855--9861.

\bibitem[Livox, 2021]{livox2021}
Livox (2021) LIO-Livox: A Robust LiDAR-Inertial Odometry for Livox LiDAR.
Available at: \url{https://github.com/Livox-SDK/LIO-Livox}.

\bibitem[Wang et al., 2021a]{wang2021lightweight}
Wang~H, Wang~C and Xie~L (2021) Lightweight 3-D localization and mapping for solid-state LiDAR.
\textit{IEEE Robotics and Automation Letters}, Vol.~6, No.~2, pp.~1801--1807.

\bibitem[Yuan et al., 2022]{yuan2022efficient}
Yuan~C, Xu~W, Liu~X, Hong~X and Zhang~F (2022) Efficient and probabilistic adaptive voxel mapping for accurate online lidar odometry.
\textit{IEEE Robotics and Automation Letters}, Vol.~7, No.~3, pp.~8518--8525.

\bibitem[Wu et al., 2023]{wu2023voxelmap++}
Wu~C, You~Y, Yuan~Y, Kong~X, Zhang~Y, Li~Q and Zhao~K (2023) Voxelmap++: Mergeable voxel mapping method for online lidar (-inertial) odometry.
\textit{IEEE Robotics and Automation Letters}, Vol.~9, No.~1, pp.~427--434.

\bibitem[Sorenson, 1966]{sorenson1966kalman}
Sorenson~HW (1966) Kalman filtering techniques.
\textit{Advances in Control Systems}, Vol.~3, pp.~219--292.

\bibitem[Cai et al., 2021]{cai2021ikd}
Cai~Y, Xu~W and Zhang~F (2021) ikd-tree: An incremental kd tree for robotic applications.
\textit{arXiv preprint arXiv:2102.10808}.

\bibitem[Chen et al., 2024a]{chen2024multi}
Chen~Q, Li~G, Xue~X and Pu~J (2024) Multi-LIO: A Lightweight Multiple LiDAR-Inertial Odometry System.
\textit{2024 IEEE International Conference on Robotics and Automation (ICRA)}, Yokohama, Japan, pp.~13748--13754.

\bibitem[Zhang et al., 2024]{zhang2024lio}
Zhang~T, Zhang~X, Liao~Z, Xia~X and Li~Y (2024) As-lio: Spatial overlap guided adaptive sliding window lidar-inertial odometry for aggressive fov variation.
\textit{2024 IEEE/RSJ International Conference on Intelligent Robots and Systems (IROS)}, Abu Dhabi, UAE, pp.~10829--10836.

\bibitem[Huang et al., 2024]{huang2024lio}
Huang~J, Zhang~Y, Xu~Q, Wu~S, Liu~J, Wang~G and Liu~W (2024) LA-LIO: Robust Localizability-Aware LiDAR-Inertial Odometry for Challenging Scenes.
\textit{2024 IEEE/RSJ International Conference on Intelligent Robots and Systems (IROS)}, Abu Dhabi, UAE, pp.~10145--10152.

\bibitem[Wang et al., 2021b]{wang2021}
Wang~H, Wang~C, Chen~C and Xie~L (2021) F-LOAM : Fast LiDAR Odometry and Mapping.
\textit{2021 IEEE/RSJ International Conference on Intelligent Robots and Systems (IROS)}, Prague, Czech Republic, pp.~4390--4396.

\bibitem[Agarwal et al., 2022]{Agarwal_Ceres_Solver_2022}
Agarwal~S, Mierle~K and The Ceres Solver Team (2022) Ceres Solver, Version~2.1.
Available at: \url{https://github.com/ceres-solver/ceres-solver}.

\bibitem[Lin and Zhang, 2024]{lin2024r}
Lin~J and Zhang~F (2024) R$^{3}$3LIVE++: A Robust, Real-Time, Radiance Reconstruction Package With a Tightly-Coupled LiDAR-Inertial-Visual State Estimator.
\textit{IEEE Transactions on Pattern Analysis and Machine Intelligence}, Vol.~46, No.~12, pp.~11168--11185.

\bibitem[Yuan et al., 2025a]{yuan2025dynamic}
Yuan~Z, Wang~X, Wu~J, Cheng~J and Yang~X (2025) LiDAR-Inertial Odometry in Dynamic Driving Scenarios using Label Consistency Detection.
\textit{2025 IEEE/RSJ International Conference on Intelligent Robots and Systems (IROS)}, Hangzhou, China, pp.~1598--1605.

\bibitem[Engel et al., 2017]{engel2017direct}
Engel~J, Koltun~V and Cremers~D (2017) Direct sparse odometry.
\textit{IEEE Transactions on Pattern Analysis and Machine Intelligence}, Vol.~40, No.~3, pp.~611--625.

\bibitem[Vizzo et al., 2023]{vizzo2023kiss}
Vizzo~I, Guadagnino~T, Mersch~B, Wiesmann~L, Behley~J and Stachniss~C (2023) Kiss-icp: In defense of point-to-point icp--simple, accurate, and robust registration if done the right way.
\textit{IEEE Robotics and Automation Letters}, Vol.~8, No.~2, pp.~1029--1036.

\bibitem[Ferrari et al., 2024]{ferrari2024mad}
Ferrari~S, Di Giammarino~L, Brizi~L and Grisetti~G (2024) MAD-ICP: It is all about matching data--robust and informed LiDAR odometry.
\textit{IEEE Robotics and Automation Letters}, Vol.~9, No.~8, pp.~7067--7074.

\bibitem[Lee et al., 2024]{lee2024genz}
Lee~D, Lim~H and Han~S (2024) GenZ-ICP: Generalizable and degeneracy-robust LiDAR odometry using an adaptive weighting.
\textit{IEEE Robotics and Automation Letters}, Vol.~9, No.~8, pp.~7118--7125.

\end{thebibliography}
\end{document}